\pdfoutput=1

\documentclass[11pt]{article}
\usepackage{booktabs}  
\usepackage{multirow}
\usepackage{acl}
\usepackage{listings} 
\usepackage{times}
\usepackage{latexsym}
\usepackage{enumerate}
\usepackage[inline]{enumitem}

\usepackage{booktabs}
\usepackage{siunitx}
\usepackage{xcolor}

\usepackage[T1]{fontenc}

\usepackage[utf8]{inputenc}

\usepackage{microtype}

\usepackage{inconsolata}

\usepackage{graphicx}
\usepackage{amsmath}
\usepackage{amssymb}

\title{Explaining Sources of Uncertainty in Automated Fact-Checking}

\author{
Jingyi Sun\thanks{Equal contribution.}~
Greta Warren\footnotemark[1]~
\textbf{
Irina Shklovski ~ 
Isabelle Augenstein}\vspace{3.5pt}
\smallskip
\\
University of Copenhagen \\
\texttt{\{jisu, grwa, ias, augenstein\}@di.ku.dk}\\
}

\usepackage{colortbl}

\newcommand{\modelblocksep}{\addlinespace[2pt]\cmidrule(lr){1-4}\addlinespace[2pt]}

\newcommand{\modelname}{CLUE} 
\newcommand{\modelvariantone}{CLUE-Span} 
\newcommand{\modelvarianttwo}{CLUE-Span+Steering} 
\newcommand{\baselinemodelname}{Prompt\textsubscript{Baseline}}
\newcommand{\modelvariantoneshortened}{CLUE-S} 
\newcommand{\modelvarianttwoshortened}{CLUE-Steer} 
\newcommand{\baselinemodelnameshortened}{Prompt\textsubscript{Base}}
\newcommand{\ModelName}{\textsc{CLUE}} 

\usepackage[T1]{fontenc}
\usepackage[utf8]{inputenc}

\begin{document}
\maketitle
\begin{abstract}

Human-AI collaboration in knowledge-intensive tasks such as fact-checking requires understanding model uncertainty in multi‑document reasoning amid conflicting/agreeing evidence. Yet, existing methods only express uncertainty as numbers or hedges without revealing which evidence conflicts cause the uncertainty, leaving users unable to resolve disagreements. 
We present \ModelName\ (\textbf{C}onflict-\&Agreement-aware \textbf{L}anguage-model \textbf{U}ncertainty \textbf{E}xplanations), a plug‑and‑play white‑box framework that, to our knowledge, is the first to generate natural‑language explanations of model uncertainty grounded in conflicting/agreeing evidence. \ModelName\ (i) identifies span‑level claim-evidence and inter‑evidence relations that signal conflict or agreement without supervision, and (ii) uses these relations to steer explanation generation, articulating how they drive the model's uncertainty. Across three language models and two fact‑checking datasets, \ModelName\ produces explanations that more faithfully track model uncertainty and better align with the model’s fact‑checking decisions than span‑agnostic explanation prompting; human raters also judge them more helpful, more informative, less redundant, and more logically consistent with the input. By explicitly tying uncertainty to evidence conflicts and agreements, \ModelName\ supports practical fact‑checking and other tasks that require reasoning over complex, conflicting information.

\end{abstract}

\section{Introduction}

Large Language Models (LLMs) are increasingly prevalent in supporting high-stakes tasks that involve reasoning about information reliability, such as fact-checking \citep{wang-etal-2024-factcheck,fontana2025evaluatingopensourcelargelanguage}.
To foster effective use of such models in fact-checking tasks, these models must explain the rationale for their predictions \citep{atanasova-etal-2020-generating-fact,kotonya_explainable_2020}.
\begin{figure}[t]
    \centering
    \includegraphics[width=\linewidth]{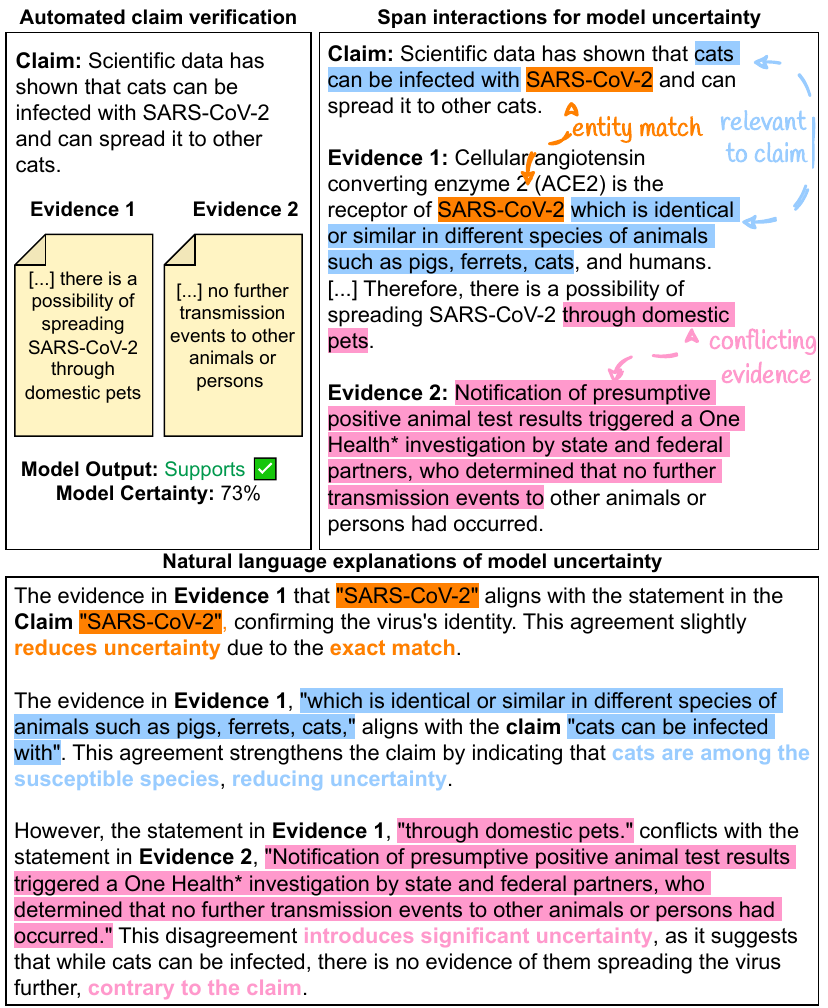}

    \caption{An example of \modelname{} output for a claim with two evidence passages: model verdict and certainty, and an uncertainty explanation grounded in conflict/agreement span interactions.}

    \label{fig:overview-example}
\end{figure}

\begin{figure*}
    \centering
        \includegraphics[width=\linewidth]{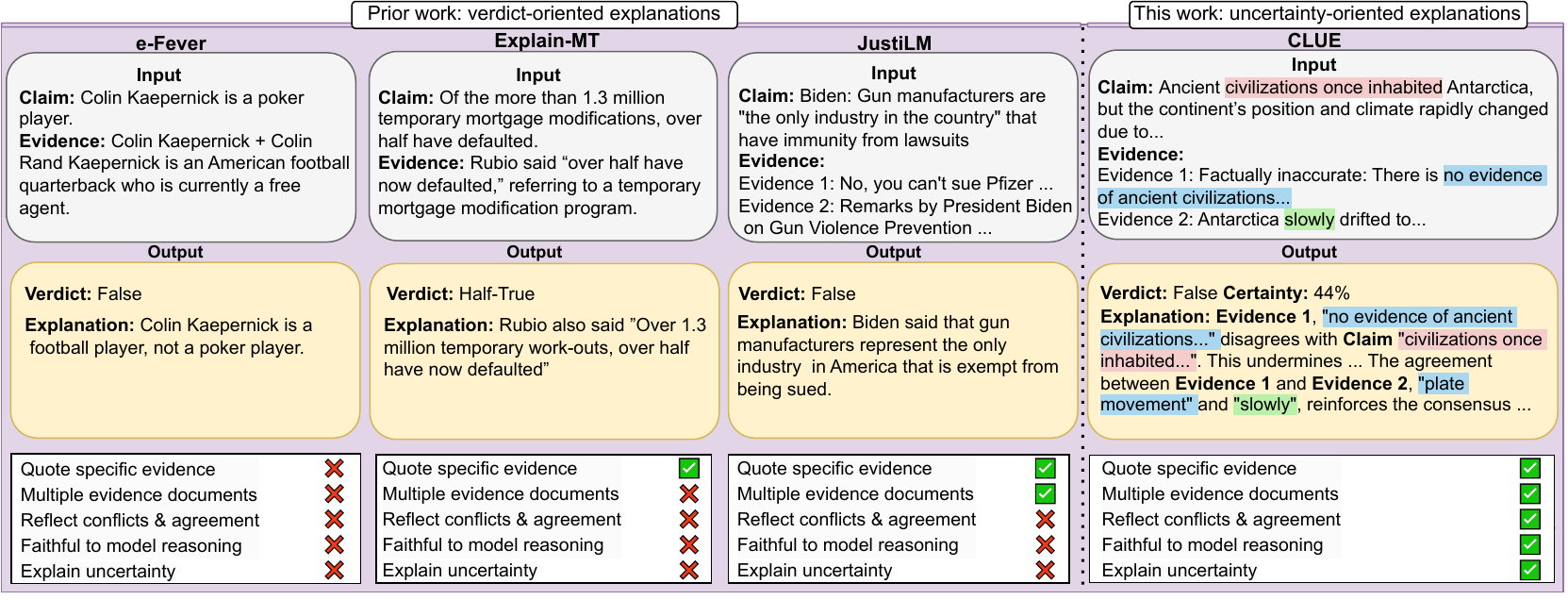}

\caption{Illustrative comparison of verdict-oriented fact-checking explanations
(e-FEVER~\citep{stammbach2020fever}, Explain-MT~\citep{atanasova-etal-2020-generating-fact}, JustiLM~\citep{zeng-gao-2024-justilm})
and CLUE's uncertainty-oriented explanations. In this comparison, CLUE is the only approach that explicitly traces predictive uncertainty to claim-evidence and inter-evidence conflict/agreement across multiple evidence passages.}



    \label{fig:expl_uncertainty}
\end{figure*}

However, current methods in automated fact-checking have been criticised for misaligning with the practical needs of fact-checkers, who must assess source reliability and reconcile conflicting evidence, rather than merely predicting a verdict label \citep{warren2025explainablefactchecking,schlichtkrull-etal-2023-intended}. Current approaches often produce explanations to justify the verdict prediction
~\citep{atanasova-etal-2020-generating-fact,stammbach2020fever,zeng-gao-2024-justilm}, however, they provide no support for communicating model uncertainty or surfacing agreements and conflicts within the evidence (See Fig.~\ref{fig:expl_uncertainty}). Methods that do express uncertainty do so via numeric scores (e.g., ``I am 73\% confident'') that are hard to contextualise \citep{zimmer1983verbal,wallsten1993preferences,van2020interpretable,liu2020intuitive} or via natural language expressions (e.g., ``I’m not sure'') that often fail to faithfully reflect model uncertainty and can inflate perceived confidence \citep{steyvers2024calibrationgapmodelhuman,yona-etal-2024-large,kim2024LLMuncertainty}. Therefore, existing explainable fact-checking systems usually exhibit two limitations: (1) neglecting to communicate model uncertainty with explanations and (2) failing to surface evidentiary conflicts and agreements that contribute to model uncertainty. This constitutes a fundamental methodological gap, as effective fact-checking requires precisely identifying the sources of uncertainty, for example, from conflicting evidence, to guide targeted verification \cite{graves2017anatomy,micallef2022factcheckers}.

To bridge this gap, we propose \ModelName, a pipeline that generates natural language explanations (NLEs) of model uncertainty by explicitly capturing conflicts and agreements in the input (e.g., a claim and its supporting or refuting evidence), see Fig. \ref{fig:overview-example}. The pipeline first identifies the salient span-level interactions that are essential to the prediction of the model through an unsupervised approach, providing an input-feature explanation that highlights key relationships between separate input segments, e.g., claim and evidence \cite{choudhury2023explaining}. Previous work has shown that these interactions are faithful to the model and plausible to humans \cite{sun2025evaluating,sun2025evaluation}. \ModelName\ then converts these signals into uncertainty-aware explanations by explicitly referring to the interactions, the conflict/agreement relations they express, and how they contribute to uncertainty regarding the verdict. \ModelName\ does not require gold-label explanations or fine-tuning, and operates entirely at inference time. 

Across three language models (\S\ref{sec:experiment_setup_models}) and two fact-checking datasets (\S\ref{sec:datasets}), we evaluate two variants of \modelname, which leverage distinct methods to guide NLE generation based on the extracted key conflicts/agreements (\S\ref{sec:nle_generation:method}). 
Automatic evaluation shows that both variants generate explanations that are more faithful to each model’s uncertainty and agree more closely with the gold fact-checking labels than a prompting baseline that lacks conflict-/agreement-span guidance (\S\ref{sec:automatic_results}). Human study participants likewise judge \modelname\ explanations as more helpful, more informative, less redundant, and more logically consistent with the input. We also observe a trade-off between two variants of our \modelname\ framework, one attains higher faithfulness, the other higher plausibility, highlighting a promising avenue for future work to achieve both simultaneously (\S\ref{sec:automatic_results}).

\section{Method}
\subsection{Preliminaries and Overall Framework}
\label{sec:framework}

Our objective is to \emph{explain why} a LLM is uncertain about a multi-evidence fact-checking instance by grounding that uncertainty in specific agreements or conflicts within the input.

\paragraph{Problem setup.}
Each input instance is a triple $X=(C,E_1,E_2)$ consisting of a claim $C$ and two evidence pieces $E_1,E_2$. In this work, we explore the scenarios in which one claim and two or three evidence pieces are presented.\footnote{As our framework is based on the pairwise input part interaction, it is readily applicable to scenarios where more claims and evidence pieces are involved.}
For clarity, we denote their concatenation as $X=[x_1,\dots,x_{|C|+|E_1|+|E_2|}]$. The task label comes from the set
$\mathcal{Y}=\{\textsc{Supports},\textsc{Refutes},\textsc{Neutral}\}$.

\paragraph{Pipeline overview.}
Our framework comprises three stages:
\begin{enumerate}[leftmargin=1.25em, itemsep=3pt,noitemsep]
    \item \textbf{Uncertainty scoring.}  
    We compute \emph{predictive entropy} from the model’s answer logits to obtain a scalar uncertainty score $u(X)$ (\S~\ref{sec:uncertainty_score_calculation}).  
    This logit-based measure is model-agnostic.

    \item \textbf{Conflicts/Agreement extraction.}  
    We capture the agreements and conflicts most relevant to the model’s reasoning by identifying the text-span interactions between $C$, $E_1$, and $E_2$ that embody these relations (\S~\ref{sec:conflicts_extraction:method}).

    \item \textbf{Explanation generation.}  
    The model receives the extracted spans as soft constraints and produces a natural language explanation $Y_R=[y_1', \dots, y_r']$ along with its predicted label $\hat{y}$ to the identified interactions (\S~\ref{sec:nle_generation:method}).  
\end{enumerate}

\paragraph{Outputs.}
For each instance $X$, the framework returns the predicted task label $\hat{y}\in\mathcal{Y}$; the numeric uncertainty score $u(X)$; and the textual explanation $Y_R=[y_1',\dots,y_r']$ that grounds the source of uncertainty in the specific agreements or conflicts between $C,E_1,E_2$.

\subsection{Predictive Uncertainty Score Generation}
\label{sec:uncertainty_score_calculation}
To quantify model uncertainty for generating an answer label on a specific input sequence, we follow previous work~\citep{kadavath2022language,yang-etal-2025-maqa} and calculate predictive uncertainty with entropy theory, which does not require multiple runs and is widely used in open-source models.

Specifically, we define the numeric uncertainty score $u$ as the entropy of the softmax distribution over the model's output logits for a set of candidate answers $\mathcal{Y} = \{\textsc{Supports}, \textsc{Refutes}, \textsc{Neutral}\}$. For each candidate label $y_i \in \mathcal{Y}$:

\begin{align}
P(y_i \mid X) &= \frac{\exp(\mathrm{logit}(y_i))}{\sum_{j=1}^{|\mathcal{Y}|} \exp(\mathrm{logit}(y_j))}
\end{align}
where $\mathrm{logit}(y_i)$ is the model's output logit towards candidate answer $y_i$ given input $X$. $P(y_i \mid X)$ is the confidence score of model for selecting $y_i$ as the final answer across all candidate answers within $\mathcal{Y}$.
Finally, the model's uncertainty towards the input sequence $X$ is:
\begin{equation}
u(X)= - \sum_{y_i \in \mathcal{Y}} P(y_i \mid X) \log P(y_i \mid X)
\label{eq:entropy_uncertainty}
\end{equation}

\subsection{Conflict and Agreement Span Interaction Identification for Answer Uncertainty}
\label{sec:conflicts_extraction:method}

To surface the conflicts and agreements that drive a model’s uncertainty, we extract and then label salient span interactions among the claim $C$ and two evidence passages, $E_1$ and $E_2$.

\paragraph{Span interaction extraction.} For each ordered input part pair $(F,T) \in \{(C,E_1), (C,E_2), (E_1,E_2)\}$, we follow previous work \cite{choudhury2023explaining,sun2025evaluating} to extract the important span interactions and their importance score to  model's answer by (i) identifying the most important attention head to the model's answer prediction from its final layer,  (ii) obtaining its attention matrix $\mathbf A \in \mathbb R^{(|F|+|T|)\times(|F|+|T|)}$, and  
(iii) symmetrizing the cross-part score $a'_{p,q}$ for each possible token interaction $(x_p,x_q)$ within $(F,T)$:
$$
a'_{p,q}= \tfrac12\bigl(\mathbf A_{p,q}+\mathbf A_{q,p}\bigr),
\quad
x_p\!\in\!F,\;x_q\!\in\!T.
$$
Based on these token interactions, we treat each $a'_{p,q}$ as an edge weight in a bipartite token graph between the two input parts $(F,T)$. Then the Louvain algorithm~\citep{blondel2008fast} is applied to search for communities of tokens with dense intra-cluster and sparse inter-cluster relationships. From each detected token community, one span interaction, $(\text{span}_w,\text{span}_v)$, can be extracted by treating the neighboring tokens of the same input part, $F$ and $T$, respectively, as a span.
By averaging the token importance score $a'_{p,q}$ of each token interaction within $(\text{span}_w,\text{span}_v)$, its importance score $a_{wv}$ can be obtained by
\begin{equation}
a_{wv}= \frac{1}{|\text{span}_w|\,|\text{span}_v|}
        \sum_{x_p\in\text{span}_w}\!
        \sum_{x_q\in\text{span}_v}\!
        a'_{p,q}.
\label{eq:span-score}
\end{equation}
The scored interactions for $(F,T)$ form
$S^{(F,T)}=\{((\text{span}_w,\text{span}_v),\,a_{wv})\}$.

\paragraph{Relation labeling.}
To tag each span interaction as \textit{agreement}, \textit{disagreement}, or \textit{unrelated},
we use a relation labeler $L_{\text{rel}}$ that can be instantiated with either an open-weight or closed LLM.
$L_{\text{rel}}$ takes the full $(C, E_1, E_2)$ context together with a candidate span pair and outputs one label in
$\{\textit{agree}, \textit{disagree}, \textit{unrelated}\}$.
We implement this using an instruction prompt (see App.~\ref{app:relation_prompt}) and report the labeling model in Sec.~\ref{sec:experiment_setup_models}.



%
After labelling all three pairs, the complete interaction set for instance $X$ is
\begin{equation}
S_R =
S_R^{(C,E_1)}\;\cup\;S_R^{(C,E_2)}\;\cup\;S_R^{(E_1,E_2)},
\end{equation}
where, for example,
$S_R^{(C,E_1)}=\{((\text{span}_w,\text{span}_v),\,a_{wv},\,r_{wv})\}$.  
Each element links two spans with an importance score and a relation label, thereby supplying the conflict- or agreement-span interactions used in later stages.

\subsection{Uncertainty Natural Language Explanation Generation}
\label{sec:nle_generation:method}

To convert the extracted conflict- and agreement-span interactions into explanations for model uncertainty, we rely on two complementary mechanisms.  
(i) \textbf{Instruction-driven prompting} embeds the spans directly in the input so the model is instructed which segments to reference.  
(ii) \textbf{Intrinsic attention steering} guides the model's own attention toward those same segments while it is generating the NLE.  
Both mechanisms use \emph{self-rationalization}: the model first states its verdict $\hat y$ and then explains $Y_R$, a sequencing shown to improve faithfulness over pipeline approaches~\citep{wiegreffe-etal-2021-measuring,marasovic-etal-2022-shot,siegel2025faithfulness}.

\paragraph{Instruction-driven NLE.}
For each instance $X$, we rank all labelled interactions by descending importance scores and keep the top $k=3$, denoted $S_R(k)$, to avoid overly long explanations.  
These three span pairs are slotted into a three-shot prompt (See App. \ref{app:baseline_prompt}), which instructs the model to explain how the highlighted agreements or conflicts influence its confidence. Finally, the standard transformer decoding process outputs both the predicted label $\hat y$ and the accompanying explanation $Y_R$.

\paragraph{Attention steering.}
Instead of explicit instructions, we can guide NLE generation by modifying attention on the fly following the attention steering method in PASTA~\citep{zhang-etal-2024-attend}.  
Starting from the same $S_R(k)$, we collect all token indices that fall inside any selected span,
\begin{equation}
\small
\mathcal{I}
=\bigl\{p \;:\; (\text{span}_w,\text{span}_v)\!\in\!S_R(k),\;
                 p\!\in\!\text{span}_w\cup\text{span}_v \bigr\}.
\end{equation}
For each attention head $(\ell,h)$ deemed relevant to model uncertainty, let $\mathbf A$ be its attention matrix.  We down-weight non-target tokens by $\beta$:

\begin{align}
\tilde{A}_{ij} &=
\frac{A_{ij}}{Z_i}
\begin{cases}
1      & \text{if } j\in\mathcal{I},\\
\beta  & \text{otherwise},
\end{cases}\\[4pt]
Z_i &= \sum_{j\in\mathcal{I}} A_{ij}
     + \beta \sum_{j\notin\mathcal{I}} A_{ij}.
\end{align}

All other heads remain unchanged.  Following \citet{zhang-etal-2024-attend}, we steer $|H| = 100$ heads and set $\beta=0.01$ to balance steering efficacy and prevent degeneration; see App.~\ref{app:attention_heads_steering_selection} for the head-selection procedure.
With the steered attention in place, the transformer generates $\hat y$ followed by the rationale $Y_R$, now naturally centered on the conflict- or agreement spans that drive its uncertainty.

\section{Experimental Setup}
\subsection{Datasets}
\label{sec:datasets}
We select two fact-checking datasets, one specific to the health domain, HealthVer \cite{sarrouti-etal-2021-evidence-based}, and one closer to a real-world fact-checking scenario, DRUID \cite{hagström2024realitycheckcontextutilisation}. These datasets were chosen because they provide multiple evidence pieces per claim, making them well-suited to our goal of explaining model uncertainty arising from the inter-evidence conflicts and agreements. For experiments, we select six hundred instances that consist of a claim and multiple pieces of evidence, and a golden label $y\in \{\textsc{Supports}, \textsc{Refutes}, \textsc{Neutral}\}$ from each dataset.\footnote{While DRUID has six fine-grained fact-checking labels, we merge the labels into the above three categories to balance the label categories.}

\subsection{Models}
\label{sec:experiment_setup_models}

We compare three generation strategies for NLEs of model uncertainty:

\begin{itemize}[leftmargin=*,noitemsep]
\item \textbf{\texttt{\baselinemodelname}}: A three-shot prompt baseline extending prior few-shot NLE work~\citep{stammbach2020fever,zeng-gao-2024-justilm,zhao2024pacar} by explicitly asking the model to highlight conflicting or supporting spans that shape its uncertainty (see prompt template in App. \ref{app:baseline_prompt}).

\item \textbf{\texttt{\modelvariantone}}: The instruction-driven variant of our \modelname\ method where the extracted span interactions are filled into a three-shot prompt to guide the explanation generation (\S\ref{sec:nle_generation:method}; App.~\ref{app:span_prompt}).

\item \textbf{\texttt{\modelvarianttwo}}: The attention steering variant of our \modelname\ method in which the same prompt as \texttt{\modelvariantone} is used. Additional attention steering is applied to instinctively guide the model's explanation generation toward the identified spans (\S\ref{sec:nle_generation:method}; App. \ref{app:span_prompt}).
\end{itemize}

We run the main experiments on multiple recent open-weight LLMs of comparable scale:
Qwen2.5-14B-Instruct~\citep{qwen2.5}, Gemma-2-9B-IT~\citep{gemma_arxiv_2024}, and OLMo-2-1124-13B-Instruct~\citep{olmo20242olmo2furious} and DeepSeek-R1-Distill-Qwen-14B~\citep{deepseek_r1_distill_qwen_14b}
(model cards in App.~\ref{app:resources}).
Each backbone is run through \modelname{} on a single NVIDIA A100-SXM 80GB GPU; we choose these models to balance
instruction-following/reasoning ability with inference efficiency (see cost report in App.~\ref{app:compute_cost}).

For span-interaction relation labeling, we instantiate $L_{\text{rel}}$ with GPT-4o~\citep{openai2024gpt4ocard}
to reduce relation-label noise (see App.~\ref{app:error_propagation} for error propagation analysis).
This step is optional: replacing $L_{\text{rel}}$ with the open-weight Qwen2.5-14B-Instruct yields comparable
downstream NLE quality with a small drop (App.~\ref{app:ablation_qwen_labeler});
we discuss this modularity and future directions in Limitations.



\section{Automatic Evaluation}
\subsection{Faithfulness}
\label{sec:faithfulness:method}

To assess whether the NLEs produced by \modelname\ are faithful to the model's uncertainty,
we adapt the Correlational Counterfactual Test (CCT)~\cite{siegel-etal-2024-probabilities}
and propose \emph{Entropy-CCT}. Entropy-CCT retains CCT's perturb-and-correlate design,
but replaces model's answer probability-shift (e.g., TVD) with \emph{predictive entropy change} to directly
target uncertainty, which is the object our NLEs aim to explain
 (see the App.~\ref{app:difference_from_cct} for comparison).

Following \citet{siegel-etal-2024-probabilities}, for each instance $X$ we construct $n$ perturbations $X'$ 
by inserting a random modifier token (adjective before a noun, or adverb before a verb; App.~\ref{app:perturbation_details}).
Let $u(X)$ denote the model's uncertainty score defined by Eq.~\ref{eq:entropy_uncertainty}.
We measure the perturbation's impact on uncertainty using Absolute Entropy Change (AEC):
\begin{equation}
\Delta u(X) = \left| u(X) - u(X^{\prime}) \right|.
\label{eq:aec}
\end{equation}

For each perturbation $X'$, we also record whether the inserted word appears in the generated NLE,
using its presence as a proxy for importance. This yields a binary mention flag
$m\in\{0,1\}$~\citep{siegel-etal-2024-probabilities,atanasova-etal-2023-faithfulness}.

Let $D_m$ denote the set of perturbed examples where the NLE \emph{mentions} the inserted word
and $D_{\lnot m}$ the complementary set where it does not.
We compute the point-biserial correlation $r_{\text{pb}}$~\citep{tate1954correlation}
between the continuous variable $\Delta u$ and the binary flag $m$ over all perturbations.
The Entropy-CCT statistic is:
\begin{equation}
\scriptstyle
\text{CCT}_{\text{entropy}} = r_{\text{pb}}=
\frac{\mathbb{E}_{m}[\Delta u] - \mathbb{E}_{\lnot m}[\Delta u]}{\mathrm{Std}(\Delta u)}
\cdot
\sqrt{ \frac{|D_m| \cdot |D_{\lnot m}|}{(|D_m| + |D_{\lnot m}|)^2} }.
\label{eq:entropy-cct}
\end{equation}

Here, $\mathbb{E}_m[\Delta u]$ and $\mathbb{E}_{\lnot m}[\Delta u]$ are the mean AEC values
for the two groups, and $\mathrm{Std}(\Delta u)$ is the standard deviation across all perturbations.
Higher Entropy-CCT indicates that the NLE more often mentions tokens whose insertion
induces larger changes in predictive uncertainty, and thus is more faithful to reflect the model's uncertainty.

\subsection{Span-Coverage}
\label{sec:spancoverage:method}

An uncertainty explanation should surface \emph{all} information conveyed by the selected span interactions.  We therefore compute \textbf{Span-Coverage}: the fraction of reference interactions that are
explicitly mentioned in the generated NLE.  Let $S_{\text{NLE}}$ be the set of span interactions extracted from the explanation, and let $S_R(k)$ be the reference set supplied in the prompt (See \S\ref{sec:nle_generation:method}).  Then
\begin{equation}
\text{Span-Coverage} \;=\;
\frac{\lvert S_{\text{NLE}} \cap S_R(k) \rvert}{\lvert S_R(k) \rvert}.
\end{equation}
A higher value indicates the NLE covers a higher proportion of the information supplied by the extracted span interactions.
\subsection{Span-Extraneous}
\label{sec:spanextraneous:method}

Ideally, the explanation should mention \emph{only} the provided interactions and avoid introducing extraneous information.  We measure the proportion of mentioned interactions that
\emph{do not} belong to the reference set, denoted
\textbf{Span-Extraneous}:
\begin{equation}
\text{Span-Extraneous} \;=\;
\frac{\lvert S_{\text{NLE}} \setminus S_R(k) \rvert }
     {\lvert S_{\text{NLE}} \rvert }.
\end{equation}
A lower value indicates closer alignment with the intended span interactions.

\subsection{Label-Explanation Entailment}
\label{sec:lee:method}
We evaluate the extent to which the uncertainty explanation agrees with the model’s predicted label by formulating the task as a natural-language inference (NLI) problem. 
First, we convert the predicted label into a hypothesis using the template 
\textit{``The claim is supported by / refuted by / neutral to the evidence.''} 
The generated explanation serves as the premise. 
The resulting premise--hypothesis pair is fed to a widely used off-the-shelf language-inference model, DeBERTa-v3~\citep{he2023debertav3improvingdebertausing}. 
The Label-Explanation Entailment (LEE) score is the proportion of examples for which the NLI model predicts \textsc{entailment}.


\subsection{Results}
\label{sec:automatic_results}

\begin{table*}[t]
\centering
\small
\setlength{\tabcolsep}{4pt}
\renewcommand{\arraystretch}{1.05}
\begin{tabular}{llcccccccc}
\toprule
\multirow{2}{*}{\textbf{Model}} & \multirow{2}{*}{\textbf{Method}}
& \multicolumn{4}{c}{\textbf{HealthVer}} & \multicolumn{4}{c}{\textbf{DRUID}} \\
\cmidrule(lr){3-6}\cmidrule(lr){7-10}
& & \textbf{Faith} $\uparrow$ & \textbf{Cov.} $\uparrow$ & \textbf{Ext.} $\downarrow$ & \textbf{LEE.} $\uparrow$
  & \textbf{Faith.} $\uparrow$ & \textbf{Cov.} $\uparrow$ & \textbf{Ext.} $\downarrow$ & \textbf{LEE.} $\uparrow$ \\
\midrule

\multirow{3}{*}{Qwen2.5-14B}
& \textbf{\texttt{\baselinemodelname}}
& -0.028 & \multicolumn{1}{c}{\textit{n/a}} & \multicolumn{1}{c}{\textit{n/a}} & 0.74
& -0.080 & \multicolumn{1}{c}{\textit{n/a}} & \multicolumn{1}{c}{\textit{n/a}} & 0.60 \\
& \textbf{\texttt{\modelvariantone}}
& {0.006} & {0.33} & {0.68} & {0.75}
& {0.089} & {0.20} & {0.38} & \textbf{0.78} \\
& \textbf{\texttt{\modelvarianttwo}}
& \textbf{0.033} & \textbf{0.44} & \textbf{0.53} & \textbf{0.80}
& \textbf{0.102} & \textbf{0.28} & \textbf{0.20} & {0.77} \\
\midrule
\multirow{3}{*}{OLMo-2-1124-13B}
& \textbf{\texttt{\baselinemodelname}}
& -0.100 & \multicolumn{1}{c}{\textit{n/a}} & \multicolumn{1}{c}{\textit{n/a}} & 0.55
& -0.130 & \multicolumn{1}{c}{\textit{n/a}} & \multicolumn{1}{c}{\textit{n/a}} & 0.53 \\
& \textbf{\texttt{\modelvariantone}}
& {0.005} & {0.10} & {0.83} & {0.61}
& {0.014} & {0.08} & {0.79} & {0.65} \\
& \textbf{\texttt{\modelvarianttwo}}
& \textbf{0.020} & \textbf{0.23} & \textbf{0.77} & \textbf{0.68}
& \textbf{0.099} & \textbf{0.15} & \textbf{0.70} & \textbf{0.69} \\
\midrule
\multirow{3}{*}{Gemma-2-9B-IT}
& \textbf{\texttt{\baselinemodelname}}
& -0.105 & \multicolumn{1}{c}{\textit{n/a}} & \multicolumn{1}{c}{\textit{n/a}} & 0.66
& -0.120 & \multicolumn{1}{c}{\textit{n/a}} & \multicolumn{1}{c}{\textit{n/a}} & 0.57 \\
& \textbf{\texttt{\modelvariantone}}
& {0.007} & {0.34} & {0.59} & {0.82}
& {0.043} & {0.23} & \textbf{0.43} & {0.76} \\
& \textbf{\texttt{\modelvarianttwo}}
& \textbf{0.021} & \textbf{0.39} & \textbf{0.50} & \textbf{0.85}
& \textbf{0.098} & \textbf{0.30} & {0.47} & \textbf{0.81} \\

\bottomrule
\end{tabular}
\caption{Automatic uncertainty-NLE evaluation. Higher is better for Faith.(\S\ref{sec:faithfulness:method})/Cov. (\S\ref{sec:spancoverage:method})/LEE. (\S\ref{sec:lee:method}) and lower is better for Ext. (\S\ref{sec:spanextraneous:method}).
\textit{n/a} indicates Cov./Ext. are undefined for \textbf{\texttt{\baselinemodelname}} because no reference spans are provided. Best results per metric for each dataset-model pair are in bold.}
\label{tab:healthver_druid_results}
\end{table*}

Here, we present the results of our automatic NLE evaluation for the scenario where the input consists of one \textit{Claim} and two pieces of \textit{Evidence}\footnote{See experiments with three pieces of evidence in App.~\ref{app:3_evidence}}. For brevity, we refer to Qwen2.5-14B-Instruct, OLMo-2-1124-13B-Instruct, and Gemma-2-9B-it simply as Qwen, OLMo, and Gemma, respectively\footnote{See experiments on the reasoning model DeepSeek‑R1‑Distill‑Qwen‑14B in App.~\ref{app:reasoning_model} }. 

\paragraph{Faithfulness.}
We use Entropy-CCT, a point--biserial correlation $r_\text{pb}$ (Eq.~\ref{eq:entropy-cct}) bounded by $[-1,1]$
to measure the faithfulness of NLEs to the model's uncertainty (\S\ref{sec:faithfulness:method}).  
When $r_{\text{pb}}=0$, the explanation mentions high- and low-impact
perturbation words equally often; every $+0.01$ adds roughly
\emph{one percentage point (pp)} to the chance that the explanation
names a token that is \emph{truly influential for the model’s predictive uncertainty} (App.~\ref{app:faithfulness_stats}).

Table~\ref{tab:healthver_druid_results} shows that
\textbf{\textbf{\texttt{\baselinemodelname}} is \emph{non-faithful} in all six settings} 
with
all $r_{\text{pb}}$ values negative ranging from $-0.03$ to $-0.13$.  
Thus its NLEs mention truly influential tokens 3--13\,pp
\emph{less} often than uninfluential ones-the opposite of faithful
behaviour. \textbf{Both variants of our \modelname\ reverse this trend.}
Presenting span interactions in the prompt
(\textbf{\texttt{\modelvariantone}}) raises every correlation to
non-negative values and peaks at $r_{\text{pb}} = 0.089$ on the
\textsc{DRUID}--Qwen setting.  This means the explanation now mentions
about 17\,pp more often than  \textbf{\texttt{\baselinemodelname}} ($r_{\text{pb}} = -0.080$). Adding attention steering
(\textbf{\texttt{\modelvarianttwo}}) lifts the $r_{\text{pb}}$ scores to
$0.033$ on \textsc{HealthVer} and $0.102$ on
\textsc{DRUID} with Qwen model, i.e., 
net gains of +6\,pp and +18\,pp over
\textbf{\texttt{\baselinemodelname}}.  Moreover, four of the six positive
correlations produced by \textbf{\texttt{\modelvarianttwo}} are
significant at $p<0.01$ (Table~\ref{tab:faith_sig_7200} in App.~\ref{app:faithfulness_stats}), confirming that
the improvements are both substantial and statistically reliable.
\textbf{Particularly large jumps of OLMo on Druid dataset (up to $\Delta r_{\text{pb}} = +0.23
\approx +23$\,pp)} suggest that span-interaction guidance from our \modelname\ framework is most
beneficial for models that initially struggle to align explanations with
its uncertainty.

\paragraph{Other Properties} We evaluate three further properties of the generated NLEs: (i) \textbf{Span-Coverage} of extracted conflict-/agreement- span interactions (\S\ref{sec:spancoverage:method}), (ii) \textbf{Span-Extraneous}: mention of non-extracted spans (\S\ref{sec:spanextraneous:method}), and (iii) \textbf{Label-Explanation Entailment} with the generated fact-checking label (\S\ref{sec:lee:method}). As Table \ref{tab:healthver_druid_results} shows, \textbf{\textbf{\textbf{\texttt{\modelvarianttwo}}} outperforms \textbf{\texttt{\modelvariantone}} in both Span-Coverage and Span-Extraneous}, consistent with the attention steering method's effectiveness in directing the model to focus on provided spans during generation \citep{zhang-etal-2024-attend}. Absolute numbers, however, remain modest (peak Span-Coverage: .44, Span-Extraneous: .20 with Qwen). A Span-Coverage of 1 means the NLE cites every extracted interaction, while a Span-Extraneous score of 0 means it adds none beyond them.
This gap highlights considerable headroom for better integrating critical span interactions into the explanations. Among the three backbones, \textbf{Qwen attains the highest Span-Coverage and the lowest Span-Extraneous scores}, a trend that likely reflects its stronger instruction-following ability (see benchmark scores in App.~\ref{app:llm_perf}), 
and thus larger or more capable models might further narrow the gap. \textbf{Both variants of our framework achieve stronger label-explanation entailment scores than the baseline}, yielding explanations logically consistent with the predicted labels while remaining faithful to the model’s uncertainty.

\section{Human Evaluation}

\subsection{Method}
We recruited N=12 participants from Prolific (\url{https://www.prolific.com/}) to evaluate 120 explanations generated by \textbf{\texttt{\baselinemodelname}}, \textbf{\texttt{\modelvariantone}}, \textbf{\texttt{\modelvarianttwo}} for 40 unique claims (20 from DRUID, 20 from HealthVer) (see details of participants and setup in App. \ref{app:participants_and_materials}).
Adapting \citet{atanasova-etal-2020-generating-fact}, participants ranked explanations in descending order (1$^{st}$, 2$^{nd}$, 3$^{rd}$) according to five criteria, complementary to our automatic evaluation metrics:
\begin{itemize}[noitemsep,leftmargin=*]
    \item \textbf{Helpfulness.} The explanation offers information that aids readers to fact-check the claim. 

    \item \textbf{Coverage.} The explanation captures \emph{all} salient information in the input that matters for the fact check, distinct from Span-Coverage (\S\ref{sec:spancoverage:method}), which counts overlap with pre-extracted spans.
    
    \item \textbf{Non-redundancy.} The explanation does not offer irrelevant or repetitive information to the input, distinct from Span-Extraneous (\S\ref{sec:spanextraneous:method}) which counts mentions outside the extracted spans.
    
    \item \textbf{Consistency.} The explanation contains logically consistent statements to the input, distinct from Label-Explanation Entailment (\S\ref{sec:lee:method}), which measures label-explanation alignment.
    
    \item \textbf{Overall Quality.} Ranking of explanations by considering all criteria above.

\end{itemize}

\subsection{Results}
\label{sec:human_eval_results}
Our evaluation shows that explanations generated by \modelname\ are consistently preferred to \texttt{\baselinemodelname}
(Table \ref{tab:human_eval}). Similar to prior work that has evaluated NLEs \cite{atanasova-etal-2020-generating-fact,huang-etal-2024-chatgpt,solano-etal-2024-sparsefit}, annotator agreement was moderate to low (App. \ref{app:IAA}), which we attribute to the 
complexity of the task.

\begin{table}[t]
    \centering
    \begin{tabular}{lcc}
    \toprule
     & \textbf{\texttt{\baselinemodelname}} & \textbf{\texttt{\modelname}} \\
     \midrule
    \textbf{Helpfulness}     &  & \\
    Overall     & 0.281  & \textbf{0.73} \\
    DRUID     &  0.312 & \textbf{0.688 }\\
    HealthVer     & 0.25  & \textbf{0.772} \\
    \\
    \textbf{Consistency}     &  & \\
    Overall     & 0.290 & \textbf{0.721}\\
    DRUID     & 0.309 & \textbf{0.691}\\
    HealthVer     & 0.27  & \textbf{0.751} \\
    \\
    \textbf{Non-redundancy}     &  & \\
    Overall     & 0.252  & \textbf{0.762} \\
    DRUID     & 0.261 & \textbf{0.739} \\
    HealthVer     & 0.242  & \textbf{0.784}\\
    \\
    \textbf{Coverage}     &  & \\
    Overall     & 0.275 & \textbf{0.722}\\
    DRUID     & 0.283 & \textbf{0.717}\\
    HealthVer     & 0.266 & \textbf{0.727} \\
    \\
    \textbf{Overall Quality}     &  & \\
    Overall     & 0.325 & \textbf{0.678}\\
    DRUID     & 0.313 & \textbf{0.688}\\
    HealthVer     & 0.336 & \textbf{0.667} \\
    \bottomrule
    \end{tabular}
    \caption{Proportion of times explanations were ranked as participants' first preference for each evaluation metric. Explanations generated by \textbf{Qwen2.5-14B-Instruct} 
    (chosen for their high faithfulness; see §\ref{sec:automatic_results})}
    \label{tab:human_eval}
\end{table}

The explanations generated by our \texttt{\modelname} framework are ranked as participants' first preference 67-78\% of the time, compared to 24-34\% of the time for \texttt{\baselinemodelname}. The \textbf{explanations generated by \textbf{\texttt{\modelname}} are rated as most helpful, containing the least amount of redundant information, highest coverage, consistency, and overall quality approximately twice as often as those generated using \texttt{\baselinemodelname}}.


Although both \texttt{\modelvariantone} and \texttt{\modelvarianttwo} outperform \texttt{\baselinemodelname} in automatic metrics (\S\ref{sec:automatic_results}) and are ranked higher than \texttt{\baselinemodelname} in the human study (Table~\ref{tab:human_eval_ranks} in App.~\ref{app:CLUE_variants}),
we observe a mild \emph{faithfulness-plausibility} trade-off when comparing the two CLUE variants. \textbf{\texttt{\modelvarianttwo}} achieves the highest automatic faithfulness, whereas \textbf{\texttt{\modelvariantone}} is ranked slightly higher in human-perceived overall quality. A possible reason is that \textbf{\texttt{\modelvarianttwo}} adheres closely to the top-$k{=}3$ span interactions (higher Span-Coverage and lower Span-Extraneous), which improves faithfulness but can reduce fluency/naturalness when some extracted spans are fragmentary. In contrast, \textbf{\texttt{\modelvariantone}} may capture additional points that participants deemed important, but not captured by the extracted span interactions~\citep{choudhury2023explaining}. This pattern aligns with the well-documented trade-off between faithfulness and plausibility~\citep{agarwal2024faithfulnessvsplausibilityunreliability,atanasova-etal-2023-faithfulness,lu2024does}, and calls for future work to enhance both aspects for improved NLE generation, e.g., by improving the quality of the extracted span interactions.


\section{Related Work}
\subsection{Uncertainty Quantification in LLMs}
Recent work estimates LLM uncertainty primarily with white-box, logit-derived measures: predictive entropy of answer distributions \citep{kadavath2022language,yang-etal-2025-maqa}, aggregation across generations \citep{malinin-gales-2021-uncertainty}, and response consistency via semantic similarity \citep{duan2024shifting,kuhn2023semantic,nikitin2024kernel}. For closed-source models, uncertainty is typically elicited through verbalised confidence \citep{lin2022tmlr_uncertainty,mielke-etal-2022-reducing}, which is overconfident and unreliable \citep{yona-etal-2024-large,tanneru2024uncertaintyLLMs}, or approximated by diversity under paraphrased prompts \citep{zhang-etal-2024-luq,chen2024quantifying}, which is compute-intensive and conflates prompt noise with model uncertainty. Accordingly, our method focuses on open-source models and adopts \emph{predictive entropy}, a simple, interpretable, efficient white-box metric computed from answer logits, avoiding prompt-induced noise.

\subsection{Linguistic Expressions of Uncertainty}
Numerical uncertainty estimates are difficult for end-users to act upon as they do not address sources of uncertainty  \citep{warren2025explainablefactchecking}. Linguistic expressions of uncertainty, e.g., phrases such as ``I'm sure''~\citep{mielke-etal-2022-reducing,tian2023just,xiong2023can,ji2025calibrating,farquhar2024detecting,kim2024LLMuncertainty}, 
may be more intuitive to understand than numerical ones \citep{zimmer1983verbal,wallsten1993preferences,windschitl1996measuring}.
However, they are not necessarily faithful reflections of the model's uncertainty \citep{yona-etal-2024-large,tanneru2024uncertaintyLLMs}
risking misleading users \citep{steyvers2024calibrationgapmodelhuman}. Moreover, they do not explain \emph{why} the model is uncertain, as our CLUE method does.

\subsection{Generating Natural Language Explanations for Fact-Checking}
Natural language explanations (NLEs) justify model predictions for lay readers in fact-checking \citep{wei-jie-etal-2024-interpretable}, complementing summarization and social-media work that provides graphs, manipulated spans as rationales \citep{ribeiro-etal-2022-factgraph,chan-etal-2023-interpretable,huang-etal-2025-manitweet,sun2025evaluation}.
In fact-checking, early systems extracted key sentences from supplied articles as explanations~\citep{atanasova-etal-2020-generating-fact}; later work improved NLE fluency via post-editing \citep{jolly2022postediting} and on data from fact-checking websites~\citep{feher2025learning,raffel2020exploring,beltagy2020longformer}. More recent methods use highlight-based explanations as signals to enhance NLE quality, employing techniques such as graph-based modeling~\citep{yuan-etal-2025-graph} and self-refinement~\citep{wang-atanasova-2025-self}. Another line of work focuses on few-shot approaches that avoid model supervision, including prompting GPT-3 to produce evidence summaries \citep{brown2020language,stammbach2020fever}, inserting a planning step~\citep{zhao2024pacar}, and leveraging retrieval-augmented language models~\citep{zeng-gao-2024-justilm}. However, existing methods are often either unfaithful to model processes~\citep{atanasova-etal-2023-faithfulness,siegel-etal-2024-probabilities,siegel2025faithfulness} or fail to address model uncertainty~\citep{warren2025explainablefactchecking}, limiting their practical utility in fact-checking~\citep{schmitt2024xai}. Recent work in a fact-checking context has shown that explanations of model uncertainty were judged to be more helpful than providing the model verdict and numerical uncertainty alone~\citep{warren2026evidenceevaluatingroleevidence}, suggesting that generating such explanations may assist people to reason about the reliability of automated fact-checking predictions.
Our framework addresses these gaps by explicitly explaining sources of uncertainty in automatic fact-checking, yielding explanations that are both faithful to model uncertainty and provide actionable insights for fact-checkers.

\section{Conclusion}
We present the first framework, \modelname, for generating NLEs of model uncertainty by referring to the conflicts and agreements between claims and multiple pieces of evidence in a fact-checking task. Our method, evaluated across three language models and two datasets, demonstrates significant improvements in both faithfulness to model uncertainty and label consistency compared to standard prompting. 
Evaluations by human participants further demonstrate that the explanations generated by \modelname\ are more helpful, more informative, less redundant, and more logically consistent with the input.
This work establishes a foundation for explainable fact-checking systems, providing end users
with grounded, faithful explanations that reflect the model's uncertainty.
Furthermore, our approach may prove useful for future work in a wide range of information-seeking and retrieval-augmented tasks (e.g., question answering) in which explaining uncertainty given conflicting context is critical.

\section*{Limitations}
\label{sec:limitations}

Our paper proposes a novel framework for generating NLEs towards the model's uncertainty by explicitly pointing to the conflicts or agreements between the claim and multiple pieces of evidence. While our framework demonstrates improved explanation quality through rigorous evaluation across multiple language models and datasets, we acknowledge several limitations. 


Our experiments are constrained to medium-sized models (Qwen2.5-14B-Instruct, Gemma2-9B-It, OLMo‑2‑1124‑13B‑Instruct, and DeepSeek‑R1‑Distill‑Qwen‑14B), which are selected based on computational limitations. Although these models show significant improvements over baseline performance, our results suggest that larger models (e.g., 70B parameter scale) may further enhance explanation quality, particularly in terms of coverage and redundancy. Our framework's modular design readily accommodates such scaling.

\modelname{} includes an auxiliary relation labeler $L_{\mathrm{rel}}$ that assigns each extracted span interaction one of \{\textsc{agree}, \textsc{disagree}, \textsc{unrelated}\}. In the main experiments we use GPT-4o for this step to reduce relation-label noise, but this step is optional: $L_{\mathrm{rel}}$ is modular, and a fully open-weight variant (e.g., Qwen2.5-14B-Instruct) yields comparable downstream NLE quality with only a small drop
(App.~\ref{app:ablation_qwen_labeler}). 
Reducing this remaining gap
while
remaining fully open-weight
may require improving relation-label prompting (more local context, higher-quality
shots) or lightly fine-tuning an open-weight model on a compact relation-judgment set
(App.~\ref{app:error_propagation_mitigation}).


Like most multi-stage pipelines, \modelname\ can occasionally propagate upstream errors: manual inspection reveals failure modes such as fragmentary spans and occasional relation mislabels, which can in turn reduce NLE faithfulness (App.~\ref{app:error_propagation}). 
These issues could be mitigated by
encouraging more complete semantic spans during extraction (e.g., mild constraints during community detection), improving relation-label robustness (prompting or lightweight fine-tuning), and adding lightweight generation-time checks that down-weight unreliable span interactions before producing the final
explanation (App.~\ref{app:error_propagation_mitigation}).

In this study, we focus on the HealthVer and DRUID datasets, which pair claims with discrete pieces of evidence, making them ideal for studying evidence-conflict scenarios. Our experiments primarily focus on setups where one claim and two or three pieces of evidence are presented. Although we did not explore setups involving more complex claim-evidence interactions, when multiple claims and/or pieces of evidence are introduced, our framework, grounded in pairwise interactions, can seamlessly accommodate more complex scenarios by simply increasing the number of input claim/evidence pieces; no methodological changes are required.

CLUE targets settings where users need to understand \emph{why} the model is uncertain by grounding uncertainty in concrete claim-evidence and inter-evidence conflicts/agreements.
This focus is motivated by qualitative findings that professional fact-checkers want systems to ``show the work'' behind a decision, especially when evidence-based tensions drive uncertainty, so that users can decide what to verify next and when to trust the output~\citep{warren2025explainablefactchecking}. While in this paper, we conduct a human study and find that uncertainty explanations generated by our proposed \modelname{} framework are more helpful than those from the simple prompting method, we do not study whether these uncertainty-oriented explanations are preferable in every context compared to alternative methods for communicating uncertainty 
(e.g., numeric confidence~\citep{zimmer1983verbal,wallsten1993preferences,van2020interpretable,liu2020intuitive} or verbal hedges~\citep{lin2022tmlr_uncertainty,mielke-etal-2022-reducing,yona-etal-2024-large,kim2024LLMuncertainty}).
In-situ evaluation with professional fact-checkers, potentially including interface-level ablation studies (e.g., tuning the number of surfaced interactions $k$ and using compact/expandable formats), would be needed to 
study the effects of uncertainty explanations on efficiency, error detection, and calibrated trust in other contexts.

Our work is limited to the scope of explaining model uncertainty arising from evidence conflicts. While this captures a critical subset of cases, real-world uncertainty may also stem from other sources, including insufficient evidence, knowledge gaps in the model, and context-memory conflicts. We view this work as a foundational step toward broader research on model uncertainty explanation.

\section*{Ethical Considerations}
\label{sec:ethics}
This work concerns automated fact-checking, which aims to reduce the harm and spread of misinformation, but nevertheless has the potential for harm or misuse through model inaccuracy, hallucination, or deployment for censorship. Our current work aims to provide explanations that allow users to examine the outputs of these systems more critically, and so we do not see any immediate risks associated with it.

Our work is limited to examining claims, evidence, and explanations in English, and so our results may not be generalisable to other languages.
As the task involved complex reasoning about technical subjects, we screened our participants to be native English speakers to ensure that they could fully understand the material and increase the chances of high-quality responses (see \ref{app:participants_and_materials} for details). However, this criteria may also introduce or reinforce existing biases and limit the generalisability of our findings.
Participants were informed about the study and its aims before agreeing to provide informed consent. No personal data was collected from participants and they received fair payment for their work (approximately 9 GBP/hour).




\section*{Acknowledgments}
$\begin{array}{l}\includegraphics[width=1cm]{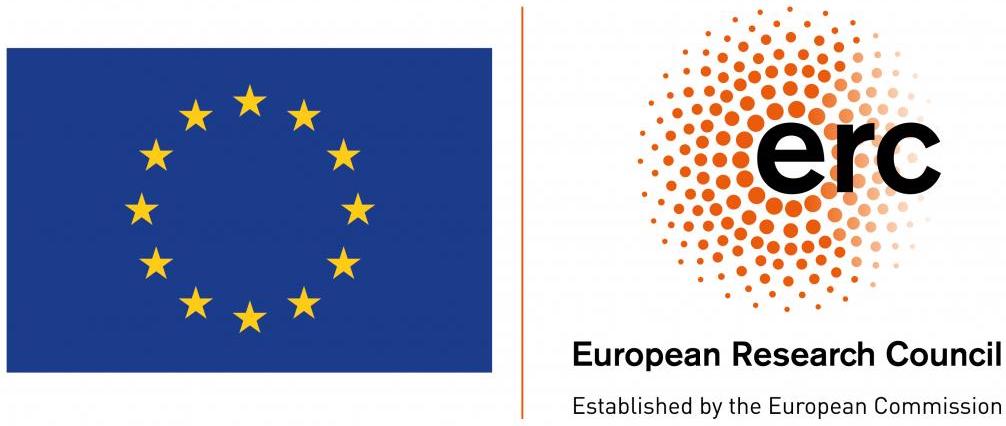} \end{array}$ 
This research was co-funded by the European Union (ERC, ExplainYourself, 101077481), by the Pioneer Centre for AI, DNRF grant number P1, as well as by The Villum Synergy Programme. Views and opinions expressed are however those of the author(s) only and do not necessarily reflect those of the European Union or the European Research Council. Neither the European Union nor the granting authority can be held responsible for them. 


\bibliography{custom_updated}

\begin{thebibliography}{76}
\providecommand{\natexlab}[1]{#1}

\bibitem[{Agarwal et~al.(2024)Agarwal, Tanneru, and Lakkaraju}]{agarwal2024faithfulnessvsplausibilityunreliability}
Chirag Agarwal, Sree~Harsha Tanneru, and Himabindu Lakkaraju. 2024.
\newblock \href {https://arxiv.org/abs/2402.04614} {{Faithfulness vs. Plausibility: On the (Un)Reliability of Explanations from Large Language Models}}.

\bibitem[{Atanasova et~al.(2023)Atanasova, Camburu, Lioma, Lukasiewicz, Simonsen, and Augenstein}]{atanasova-etal-2023-faithfulness}
Pepa Atanasova, Oana-Maria Camburu, Christina Lioma, Thomas Lukasiewicz, Jakob~Grue Simonsen, and Isabelle Augenstein. 2023.
\newblock \href {https://doi.org/10.18653/v1/2023.acl-short.25} {{Faithfulness Tests for Natural Language Explanations}}.
\newblock In \emph{Proceedings of the 61st Annual Meeting of the Association for Computational Linguistics (Volume 2: Short Papers)}, pages 283--294, Toronto, Canada. Association for Computational Linguistics.

\bibitem[{Atanasova et~al.(2020)Atanasova, Simonsen, Lioma, and Augenstein}]{atanasova-etal-2020-generating-fact}
Pepa Atanasova, Jakob~Grue Simonsen, Christina Lioma, and Isabelle Augenstein. 2020.
\newblock \href {https://doi.org/10.18653/v1/2020.acl-main.656} {{Generating Fact Checking Explanations}}.
\newblock In \emph{Proceedings of the 58th Annual Meeting of the Association for Computational Linguistics}, pages 7352--7364, Online. Association for Computational Linguistics.

\bibitem[{Beltagy et~al.(2020)Beltagy, Peters, and Cohan}]{beltagy2020longformer}
Iz~Beltagy, Matthew~E. Peters, and Arman Cohan. 2020.
\newblock \href {https://arxiv.org/abs/2004.05150} {{Longformer: The Long-Document Transformer}}.
\newblock \emph{ArXiv preprint}, abs/2004.05150.

\bibitem[{Bird et~al.(2009)Bird, Klein, and Loper}]{bird_klein_loper_2009_nltk}
Steven Bird, Ewan Klein, and Edward Loper. 2009.
\newblock \emph{Natural Language Processing with Python}.
\newblock O'Reilly Media.

\bibitem[{Blondel et~al.(2008)Blondel, Guillaume, Lambiotte, and Lefebvre}]{blondel2008fast}
Vincent~D. Blondel, Jean-Loup Guillaume, Renaud Lambiotte, and Etienne Lefebvre. 2008.
\newblock \href {https://doi.org/10.1088/1742-5468/2008/10/P10008} {{Fast Unfolding of Communities in Large Networks}}.
\newblock \emph{Journal of statistical mechanics: theory and experiment}, 2008(10):P10008.

\bibitem[{Brown et~al.(2020)Brown, Mann, Ryder, Subbiah, Kaplan, Dhariwal, Neelakantan, Shyam, Sastry, Askell, Agarwal, Herbert{-}Voss, Krueger, Henighan, Child, Ramesh, Ziegler, Wu, Winter, Hesse, Chen, Sigler, Litwin, Gray, Chess, Clark, Berner, McCandlish, Radford, Sutskever, and Amodei}]{brown2020language}
Tom~B. Brown, Benjamin Mann, Nick Ryder, Melanie Subbiah, Jared Kaplan, Prafulla Dhariwal, Arvind Neelakantan, Pranav Shyam, Girish Sastry, Amanda Askell, Sandhini Agarwal, Ariel Herbert{-}Voss, Gretchen Krueger, Tom Henighan, Rewon Child, Aditya Ramesh, Daniel~M. Ziegler, Jeffrey Wu, Clemens Winter, Christopher Hesse, Mark Chen, Eric Sigler, Mateusz Litwin, Scott Gray, Benjamin Chess, Jack Clark, Christopher Berner, Sam McCandlish, Alec Radford, Ilya Sutskever, and Dario Amodei. 2020.
\newblock \href {https://proceedings.neurips.cc/paper/2020/hash/1457c0d6bfcb4967418bfb8ac142f64a-Abstract.html} {Language models are few-shot learners}.
\newblock In \emph{Advances in Neural Information Processing Systems 33: Annual Conference on Neural Information Processing Systems 2020, NeurIPS 2020, December 6-12, 2020, virtual}.

\bibitem[{Chan et~al.(2023)Chan, Zeng, and Ji}]{chan-etal-2023-interpretable}
Hou~Pong Chan, Qi~Zeng, and Heng Ji. 2023.
\newblock \href {https://doi.org/10.18653/v1/2023.findings-acl.402} {{Interpretable Automatic Fine-grained Inconsistency Detection in Text Summarization}}.
\newblock In \emph{Findings of the Association for Computational Linguistics: ACL 2023}, pages 6433--6444, Toronto, Canada. Association for Computational Linguistics.

\bibitem[{Chen and Mueller(2024)}]{chen2024quantifying}
Jiuhai Chen and Jonas Mueller. 2024.
\newblock \href {https://doi.org/10.18653/v1/2024.acl-long.283} {{Quantifying Uncertainty in Answers from any Language Model and Enhancing their Trustworthiness}}.
\newblock In \emph{Proceedings of the 62nd Annual Meeting of the Association for Computational Linguistics (Volume 1: Long Papers)}, pages 5186--5200, Bangkok, Thailand. Association for Computational Linguistics.

\bibitem[{Cobbe et~al.(2021)Cobbe, Kosaraju, Bavarian, Chen, Jun, Kaiser, Plappert, Tworek, Hilton, Nakano et~al.}]{cobbe2021training}
Karl Cobbe, Vineet Kosaraju, Mohammad Bavarian, Mark Chen, Heewoo Jun, Lukasz Kaiser, Matthias Plappert, Jerry Tworek, Jacob Hilton, Reiichiro Nakano, et~al. 2021.
\newblock \href {https://arxiv.org/abs/2110.14168} {{Training Verifiers to Solve Math Word Problems}}.
\newblock \emph{ArXiv preprint}, abs/2110.14168.

\bibitem[{{DeepSeek-AI}()}]{deepseek_r1_distill_qwen_14b}
{DeepSeek-AI}.
\newblock {DeepSeek-R1-Distill-Qwen-14B}.
\newblock \url{https://huggingface.co/deepseek-ai/DeepSeek-R1-Distill-Qwen-14B}.
\newblock HuggingFace model card; accessed 2025-10-05.

\bibitem[{Duan et~al.(2024)Duan, Cheng, Wang, Zavalny, Wang, Xu, Kailkhura, and Xu}]{duan2024shifting}
Jinhao Duan, Hao Cheng, Shiqi Wang, Alex Zavalny, Chenan Wang, Renjing Xu, Bhavya Kailkhura, and Kaidi Xu. 2024.
\newblock \href {https://doi.org/10.18653/v1/2024.acl-long.276} {{Shifting Attention to Relevance: Towards the Predictive Uncertainty Quantification of Free-Form Large Language Models}}.
\newblock In \emph{Proceedings of the 62nd Annual Meeting of the Association for Computational Linguistics (Volume 1: Long Papers)}, pages 5050--5063, Bangkok, Thailand. Association for Computational Linguistics.

\bibitem[{Farquhar et~al.(2024)Farquhar, Kossen, Kuhn, and Gal}]{farquhar2024detecting}
Sebastian Farquhar, Jannik Kossen, Lorenz Kuhn, and Yarin Gal. 2024.
\newblock \href {https://doi.org/10.1038/s41586-024-07421-0} {{Detecting Hallucinations in Large Language Models using Semantic Entropy}}.
\newblock \emph{Nature}, 630(8017):625--630.

\bibitem[{Feher et~al.(2025)Feher, Khered, Zhang, Batista-Navarro, and Schlegel}]{feher2025learning}
Darius Feher, Abdullah Khered, Hao Zhang, Riza Batista-Navarro, and Viktor Schlegel. 2025.
\newblock \href {https://doi.org/10.1016/j.engappai.2024.109492} {{Learning to Generate and Evaluate Fact-Checking Explanations with Transformers}}.
\newblock \emph{Engineering Applications of Artificial Intelligence}, 139:109492.

\bibitem[{Fontana et~al.(2025)Fontana, Corso, Zuccolotto, and Pierri}]{fontana2025evaluatingopensourcelargelanguage}
Nicolo' Fontana, Francesco Corso, Enrico Zuccolotto, and Francesco Pierri. 2025.
\newblock \href {https://arxiv.org/abs/2503.05565} {{Evaluating Open-source Large Language Models for Automated Fact-checking}}.
\newblock \emph{ArXiv preprint}, abs/2503.05565.

\bibitem[{{Gemma Team}(2024)}]{gemma_arxiv_2024}
{Gemma Team}. 2024.
\newblock \href {https://arxiv.org/abs/2403.08295} {{Gemma: Open Models Based on Gemini Research and Technology}}.
\newblock \emph{ArXiv preprint}, abs/2403.08295.

\bibitem[{Graves(2017)}]{graves2017anatomy}
Lucas Graves. 2017.
\newblock \href {https://doi.org/10.1111/cccr.12163} {Anatomy of a fact check: Objective practice and the contested epistemology of fact checking}.
\newblock \emph{{Communication, Culture \& Critique}}, 10(3):518--537.

\bibitem[{Hagström et~al.(2024)Hagström, Marjanović, Yu, Arora, Lioma, Maistro, Atanasova, and Augenstein}]{hagström2024realitycheckcontextutilisation}
Lovisa Hagström, Sara~Vera Marjanović, Haeun Yu, Arnav Arora, Christina Lioma, Maria Maistro, Pepa Atanasova, and Isabelle Augenstein. 2024.
\newblock \href {https://arxiv.org/abs/2412.17031} {{A Reality Check on Context Utilisation for Retrieval-Augmented Generation}}.

\bibitem[{He et~al.(2023)He, Gao, and Chen}]{he2023debertav3improvingdebertausing}
Pengcheng He, Jianfeng Gao, and Weizhu Chen. 2023.
\newblock \href {https://openreview.net/pdf?id=sE7-XhLxHA} {{DeBERTaV3: Improving DeBERTa using ELECTRA-Style Pre-Training with Gradient-Disentangled Embedding Sharing}}.
\newblock In \emph{The Eleventh International Conference on Learning Representations, {ICLR} 2023, Kigali, Rwanda, May 1-5, 2023}. OpenReview.net.

\bibitem[{Hendrycks et~al.(2021)Hendrycks, Burns, Basart, Zou, Mazeika, Song, and Steinhardt}]{hendrycks2021measuring}
Dan Hendrycks, Collin Burns, Steven Basart, Andy Zou, Mantas Mazeika, Dawn Song, and Jacob Steinhardt. 2021.
\newblock \href {https://openreview.net/forum?id=d7KBjmI3GmQ} {{Measuring Massive Multitask Language Understanding}}.
\newblock In \emph{9th International Conference on Learning Representations, {ICLR} 2021, Virtual Event, Austria, May 3-7, 2021}. OpenReview.net.

\bibitem[{Honnibal and Montani(2017)}]{honnibal_montani_spacy2_2017}
Matthew Honnibal and Ines Montani. 2017.
\newblock \href {https://spacy.io/} {{spaCy 2: Natural Language Understanding with Bloom Embeddings, Convolutional Neural Networks and Incremental Parsing}}.

\bibitem[{Huang et~al.(2024)Huang, Kwak, Park, and An}]{huang-etal-2024-chatgpt}
Fan Huang, Haewoon Kwak, Kunwoo Park, and Jisun An. 2024.
\newblock \href {https://aclanthology.org/2024.lrec-main.277} {{C}hat{GPT} {Rates Natural Language Explanation Quality like Humans: But on Which Scales?}}
\newblock In \emph{Proceedings of the 2024 Joint International Conference on Computational Linguistics, Language Resources and Evaluation (LREC-COLING 2024)}, pages 3111--3132, Torino, Italia. ELRA and ICCL.

\bibitem[{Huang et~al.(2025)Huang, Chan, McKeown, and Ji}]{huang-etal-2025-manitweet}
Kung-Hsiang Huang, Hou~Pong Chan, Kathleen McKeown, and Heng Ji. 2025.
\newblock \href {https://aclanthology.org/2025.coling-main.739/} {{M}ani{T}weet: {A New Benchmark for Identifying Manipulation of News on Social Media}}.
\newblock In \emph{Proceedings of the 31st International Conference on Computational Linguistics}, pages 11161--11180, Abu Dhabi, UAE. Association for Computational Linguistics.

\bibitem[{Ji et~al.(2025)Ji, Yu, Koishekenov, Bang, Hartshorn, Schelten, Zhang, Fung, and Cancedda}]{ji2025calibrating}
Ziwei Ji, Lei Yu, Yeskendir Koishekenov, Yejin Bang, Anthony Hartshorn, Alan Schelten, Cheng Zhang, Pascale Fung, and Nicola Cancedda. 2025.
\newblock \href {https://arxiv.org/abs/2503.14477} {{Calibrating Verbal Uncertainty as a Linear Feature to Reduce Hallucinations}}.
\newblock \emph{ArXiv preprint}, abs/2503.14477.

\bibitem[{Jolly et~al.(2022)Jolly, Atanasova, and Augenstein}]{jolly2022postediting}
Shailza Jolly, Pepa Atanasova, and Isabelle Augenstein. 2022.
\newblock \href {https://doi.org/10.3390/info13100500} {Generating fluent fact checking explanations with unsupervised post-editing}.
\newblock \emph{Information}, 13(10).

\bibitem[{Kadavath et~al.(2022)Kadavath, Conerly, Askell, Henighan, Drain, Perez, Schiefer, Hatfield-Dodds, DasSarma, Tran-Johnson et~al.}]{kadavath2022language}
Saurav Kadavath, Tom Conerly, Amanda Askell, Tom Henighan, Dawn Drain, Ethan Perez, Nicholas Schiefer, Zac Hatfield-Dodds, Nova DasSarma, Eli Tran-Johnson, et~al. 2022.
\newblock \href {https://arxiv.org/abs/2207.05221} {{Language Models (Mostly) Know What They Know}}.
\newblock \emph{ArXiv preprint}, abs/2207.05221.

\bibitem[{Kendall and Smith(1939)}]{kendall1939problem}
Maurice~G Kendall and B.~Babington Smith. 1939.
\newblock \href {https://arxiv.org/abs/https://www.jstor.org/stable/2235668} {{The Problem of M Rankings}}.
\newblock \emph{The annals of mathematical statistics}, 10(3):275--287.

\bibitem[{Kim et~al.(2024)Kim, Liao, Vorvoreanu, Ballard, and Vaughan}]{kim2024LLMuncertainty}
Sunnie S.~Y. Kim, Q.~Vera Liao, Mihaela Vorvoreanu, Stephanie Ballard, and Jennifer~Wortman Vaughan. 2024.
\newblock \href {https://doi.org/10.1145/3630106.3658941} {{"I'm Not Sure, But...": Examining the Impact of Large Language Models' Uncertainty Expression on User Reliance and Trust}}.
\newblock In \emph{Proceedings of the 2024 ACM Conference on Fairness, Accountability, and Transparency}, FAccT '24, page 822–835, New York, NY, USA. Association for Computing Machinery.

\bibitem[{Kotonya and Toni(2020)}]{kotonya_explainable_2020}
Neema Kotonya and Francesca Toni. 2020.
\newblock \href {https://doi.org/10.18653/v1/2020.coling-main.474} {{Explainable Automated Fact-Checking: A Survey}}.
\newblock In \emph{Proceedings of the 28th International Conference on Computational Linguistics}, pages 5430--5443, Barcelona, Spain (Online). International Committee on Computational Linguistics.

\bibitem[{Kuhn et~al.(2023)Kuhn, Gal, and Farquhar}]{kuhn2023semantic}
Lorenz Kuhn, Yarin Gal, and Sebastian Farquhar. 2023.
\newblock \href {https://openreview.net/pdf?id=VD-AYtP0dve} {{Semantic Uncertainty: Linguistic Invariances for Uncertainty Estimation in Natural Language Generation}}.
\newblock In \emph{The Eleventh International Conference on Learning Representations, {ICLR} 2023, Kigali, Rwanda, May 1-5, 2023}. OpenReview.net.

\bibitem[{Lin et~al.(2022)Lin, Hilton, and Evans}]{lin2022tmlr_uncertainty}
Stephanie~C. Lin, Jacob Hilton, and Owain Evans. 2022.
\newblock \href {https://doi.org/10.48550/arXiv.2205.14334} {{Teaching Models to Express Their Uncertainty in Words}}.
\newblock \emph{Transactions on Machine Learning Research}.
\newblock \url{https://openreview.net/forum?id=8s8K2UZGTZ}.

\bibitem[{Liu et~al.(2020)Liu, Juanchich, Sirota, and Orbell}]{liu2020intuitive}
Dawn Liu, Marie Juanchich, Miroslav Sirota, and Sheina Orbell. 2020.
\newblock \href {https://doi.org/10.1177/1747021820903439} {{The Intuitive Use of Contextual Information in Decisions Made with Verbal and Numerical Quantifiers}}.
\newblock \emph{Quarterly Journal of Experimental Psychology}, 73(4):481--494.

\bibitem[{Lu and Ma(2024)}]{lu2024does}
Xiaolei Lu and Jianghong Ma. 2024.
\newblock \href {https://arxiv.org/abs/2404.00140} {{Does Faithfulness Conflict with Plausibility? An Empirical Study in Explainable AI across NLP Tasks}}.
\newblock \emph{ArXiv preprint}, abs/2404.00140.

\bibitem[{Malinin and Gales(2021)}]{malinin-gales-2021-uncertainty}
Andrey Malinin and Mark J.~F. Gales. 2021.
\newblock \href {https://openreview.net/forum?id=jN5y-zb5Q7m} {{Uncertainty Estimation in Autoregressive Structured Prediction}}.
\newblock In \emph{9th International Conference on Learning Representations, {ICLR} 2021, Virtual Event, Austria, May 3-7, 2021}. OpenReview.net.

\bibitem[{Marasovic et~al.(2022)Marasovic, Beltagy, Downey, and Peters}]{marasovic-etal-2022-shot}
Ana Marasovic, Iz~Beltagy, Doug Downey, and Matthew Peters. 2022.
\newblock \href {https://doi.org/10.18653/v1/2022.findings-naacl.31} {{Few-Shot Self-Rationalization with Natural Language Prompts}}.
\newblock In \emph{Findings of the Association for Computational Linguistics: NAACL 2022}, pages 410--424, Seattle, United States. Association for Computational Linguistics.

\bibitem[{Micallef et~al.(2022)Micallef, Armacost, Memon, and Patil}]{micallef2022factcheckers}
Nicholas Micallef, Vivienne Armacost, Nasir Memon, and Sameer Patil. 2022.
\newblock \href {https://doi.org/10.1145/3512974} {{True or False: Studying the Work Practices of Professional Fact-Checkers}}.
\newblock \emph{Proc. ACM Hum.-Comput. Interact.}, 6(CSCW1).

\bibitem[{Mielke et~al.(2022)Mielke, Szlam, Dinan, and Boureau}]{mielke-etal-2022-reducing}
Sabrina~J. Mielke, Arthur Szlam, Emily Dinan, and Y-Lan Boureau. 2022.
\newblock \href {https://doi.org/10.1162/tacl_a_00494} {{Reducing Conversational Agents' Overconfidence Through Linguistic Calibration}}.
\newblock \emph{Transactions of the Association for Computational Linguistics}, 10:857--872.

\bibitem[{Miller(1992)}]{miller_1995_wordnet}
George~A. Miller. 1992.
\newblock \href {https://aclanthology.org/H92-1116} {{W}ord{N}et: {A Lexical Database} for {E}nglish}.
\newblock In \emph{Speech and Natural Language: Proceedings of a Workshop Held at Harriman, New York, {F}ebruary 23-26, 1992}.

\bibitem[{Nikitin et~al.(2024)Nikitin, Kossen, Gal, and Marttinen}]{nikitin2024kernel}
Alexander Nikitin, Jannik Kossen, Yarin Gal, and Pekka Marttinen. 2024.
\newblock \href {http://papers.nips.cc/paper\_files/paper/2024/hash/10c456d2160517581a234dfde15a7505-Abstract-Conference.html} {{Kernel Language Entropy: Fine-grained Uncertainty Quantification for LLMs from Semantic Similarities}}.
\newblock In \emph{Advances in Neural Information Processing Systems 38: Annual Conference on Neural Information Processing Systems 2024, NeurIPS 2024, Vancouver, BC, Canada, December 10 - 15, 2024}.

\bibitem[{{OpenAI Team}(2024)}]{openai2024gpt4ocard}
{OpenAI Team}. 2024.
\newblock \href {https://arxiv.org/abs/2410.21276} {{GPT-4o System Card}}.

\bibitem[{{Qwen Team}(2024)}]{qwen2.5}
{Qwen Team}. 2024.
\newblock \href {https://qwenlm.github.io/blog/qwen2.5/} {{Qwen2.5: A Party of Foundation Models}}.

\bibitem[{Raffel et~al.(2020)Raffel, Shazeer, Roberts, Lee, Narang, Matena, Zhou, Li, and Liu}]{raffel2020exploring}
Colin Raffel, Noam Shazeer, Adam Roberts, Katherine Lee, Sharan Narang, Michael Matena, Yanqi Zhou, Wei Li, and Peter~J. Liu. 2020.
\newblock \href {http://jmlr.org/papers/v21/20-074.html} {{Exploring the Limits of Transfer Learning with a Unified Text-to-Text Transformer}}.
\newblock \emph{J. Mach. Learn. Res.}, 21:140:1--140:67.

\bibitem[{Ray~Choudhury et~al.(2023)Ray~Choudhury, Atanasova, and Augenstein}]{choudhury2023explaining}
Sagnik Ray~Choudhury, Pepa Atanasova, and Isabelle Augenstein. 2023.
\newblock \href {https://doi.org/10.18653/v1/2023.emnlp-main.783} {{Explaining Interactions Between Text Spans}}.
\newblock In \emph{Proceedings of the 2023 Conference on Empirical Methods in Natural Language Processing}, pages 12709--12730, Singapore. Association for Computational Linguistics.

\bibitem[{Ribeiro et~al.(2022)Ribeiro, Liu, Gurevych, Dreyer, and Bansal}]{ribeiro-etal-2022-factgraph}
Leonardo F.~R. Ribeiro, Mengwen Liu, Iryna Gurevych, Markus Dreyer, and Mohit Bansal. 2022.
\newblock \href {https://doi.org/10.18653/v1/2022.naacl-main.236} {{F}act{G}raph: {Evaluating Factuality in Summarization with Semantic Graph Representations}}.
\newblock In \emph{Proceedings of the 2022 Conference of the North American Chapter of the Association for Computational Linguistics: Human Language Technologies}, pages 3238--3253, Seattle, United States. Association for Computational Linguistics.

\bibitem[{Sarrouti et~al.(2021)Sarrouti, Ben~Abacha, Mrabet, and Demner-Fushman}]{sarrouti-etal-2021-evidence-based}
Mourad Sarrouti, Asma Ben~Abacha, Yassine Mrabet, and Dina Demner-Fushman. 2021.
\newblock \href {https://doi.org/10.18653/v1/2021.findings-emnlp.297} {{Evidence-based Fact-Checking of Health-related Claims}}.
\newblock In \emph{Findings of the Association for Computational Linguistics: EMNLP 2021}, pages 3499--3512, Punta Cana, Dominican Republic. Association for Computational Linguistics.

\bibitem[{Schlichtkrull et~al.(2023)Schlichtkrull, Ousidhoum, and Vlachos}]{schlichtkrull-etal-2023-intended}
Michael Schlichtkrull, Nedjma Ousidhoum, and Andreas Vlachos. 2023.
\newblock \href {https://doi.org/10.18653/v1/2023.findings-emnlp.577} {{The Intended Uses of Automated Fact-Checking Artefacts: Why, How and Who}}.
\newblock In \emph{Findings of the Association for Computational Linguistics: EMNLP 2023}, pages 8618--8642, Singapore. Association for Computational Linguistics.

\bibitem[{Schmitt et~al.(2024)Schmitt, Villa-Arenas, Feldhus, Meyer, Spang, and M\"{o}ller}]{schmitt2024xai}
Vera Schmitt, Luis-Felipe Villa-Arenas, Nils Feldhus, Joachim Meyer, Robert~P. Spang, and Sebastian M\"{o}ller. 2024.
\newblock \href {https://doi.org/10.1145/3630106.3659031} {{The Role of Explainability in Collaborative Human-AI Disinformation Detection}}.
\newblock In \emph{Proceedings of the 2024 ACM Conference on Fairness, Accountability, and Transparency}, FAccT '24, page 2157–2174, New York, NY, USA. Association for Computing Machinery.

\bibitem[{Siegel et~al.(2024)Siegel, Camburu, Heess, and Perez-Ortiz}]{siegel-etal-2024-probabilities}
Noah Siegel, Oana-Maria Camburu, Nicolas Heess, and Maria Perez-Ortiz. 2024.
\newblock \href {https://doi.org/10.18653/v1/2024.acl-short.49} {{The Probabilities Also Matter: A More Faithful Metric for Faithfulness of Free-Text Explanations in Large Language Models}}.
\newblock In \emph{Proceedings of the 62nd Annual Meeting of the Association for Computational Linguistics (Volume 2: Short Papers)}, pages 530--546, Bangkok, Thailand. Association for Computational Linguistics.

\bibitem[{Siegel et~al.(2025)Siegel, Heess, Perez-Ortiz, and Camburu}]{siegel2025faithfulness}
Noah~Y Siegel, Nicolas Heess, Maria Perez-Ortiz, and Oana-Maria Camburu. 2025.
\newblock \href {https://arxiv.org/abs/2503.13445} {{Faithfulness of LLM Self-Explanations for Commonsense Tasks: Larger Is Better, and Instruction-Tuning Allows Trade-Offs but Not Pareto Dominance}}.
\newblock \emph{ArXiv preprint}, abs/2503.13445.

\bibitem[{Solano et~al.(2024)Solano, Sanni, Camburu, and Minervini}]{solano-etal-2024-sparsefit}
Jesus Solano, Mardhiyah Sanni, Oana-Maria Camburu, and Pasquale Minervini. 2024.
\newblock \href {https://doi.org/10.18653/v1/2024.acl-long.113} {{S}parse{F}it: {Few-shot Prompting with Sparse Fine-tuning for Jointly Generating Predictions and Natural Language Explanations}}.
\newblock In \emph{Proceedings of the 62nd Annual Meeting of the Association for Computational Linguistics (Volume 1: Long Papers)}, pages 2053--2077, Bangkok, Thailand. Association for Computational Linguistics.

\bibitem[{Stammbach and Ash(2020)}]{stammbach2020fever}
Dominik Stammbach and Elliott Ash. 2020.
\newblock \href {https://doi.org/10.3929/ethz-b-000453826} {{e-FEVER: Explanations and Summaries for Automated Fact Checking}}.
\newblock \emph{Proceedings of the 2020 Truth and Trust Online (TTO 2020)}, pages 32--43.

\bibitem[{Steyvers et~al.(2025)Steyvers, Tejeda, Kumar, Belem, Karny, Hu, Mayer, and Smyth}]{steyvers2024calibrationgapmodelhuman}
Mark Steyvers, Heliodoro Tejeda, Aakriti Kumar, Catarina Belem, Sheer Karny, Xinyue Hu, Lukas~W Mayer, and Padhraic Smyth. 2025.
\newblock \href {https://doi.org/10.1038/s42256-024-00976-7} {{What Large Language Models Know and What People Yhink They Know}}.
\newblock \emph{Nature Machine Intelligence}, pages 1--11.

\bibitem[{Sun et~al.(2025)Sun, Atanasova, and Augenstein}]{sun2025evaluating}
Jingyi Sun, Pepa Atanasova, and Isabelle Augenstein. 2025.
\newblock \href {https://aclanthology.org/2025.naacl-long.530/} {{Evaluating Input Feature Explanations through a Unified Diagnostic Evaluation Framework}}.
\newblock In \emph{Proceedings of the 2025 Conference of the Nations of the Americas Chapter of the Association for Computational Linguistics: Human Language Technologies (Volume 1: Long Papers)}, pages 10559--10577, Albuquerque, New Mexico. Association for Computational Linguistics.

\bibitem[{Sun et~al.(2026)Sun, Atanasova, Choudhury, Islam, and Augenstein}]{sun2025evaluation}
Jingyi Sun, Pepa Atanasova, Sagnik~Ray Choudhury, Sekh~Mainul Islam, and Isabelle Augenstein. 2026.
\newblock \href {https://doi.org/10.1162/COLI.a.621} {{Evaluation Framework for Highlight Explanations of Context Utilisation in Language Models}}.
\newblock \emph{Computational Linguistics}, pages 1--33.

\bibitem[{Tanneru et~al.(2024)Tanneru, Agarwal, and Lakkaraju}]{tanneru2024uncertaintyLLMs}
Sree~Harsha Tanneru, Chirag Agarwal, and Himabindu Lakkaraju. 2024.
\newblock \href {https://proceedings.mlr.press/v238/harsha-tanneru24a.html} {{Quantifying Uncertainty in Natural Language Explanations of Large Language Models}}.
\newblock In \emph{International Conference on Artificial Intelligence and Statistics, 2-4 May 2024, Palau de Congressos, Valencia, Spain}, volume 238 of \emph{Proceedings of Machine Learning Research}, pages 1072--1080. {PMLR}.

\bibitem[{Tate(1954)}]{tate1954correlation}
Robert~F Tate. 1954.
\newblock \href {https://www.jstor.org/stable/2236844?seq=1} {{Correlation between a Discrete and a Continuous Variable. Point-Biserial Correlation}}.
\newblock \emph{The Annals of mathematical statistics}, 25(3):603--607.

\bibitem[{{Team OLMo} et~al.(2025){Team OLMo}, Walsh, Soldaini, Groeneveld, Lo, Arora, Bhagia, Gu, Huang, Jordan, Lambert, Schwenk, Tafjord, Anderson, Atkinson, Brahman, Clark, Dasigi, Dziri, Guerquin, Ivison, Koh, Liu, Malik, Merrill, Miranda, Morrison, Murray, Nam, Pyatkin, Rangapur, Schmitz, Skjonsberg, Wadden, Wilhelm, Wilson, Zettlemoyer, Farhadi, Smith, and Hajishirzi}]{olmo20242olmo2furious}
{Team OLMo}, Pete Walsh, Luca Soldaini, Dirk Groeneveld, Kyle Lo, Shane Arora, Akshita Bhagia, Yuling Gu, Shengyi Huang, Matt Jordan, Nathan Lambert, Dustin Schwenk, Oyvind Tafjord, Taira Anderson, David Atkinson, Faeze Brahman, Christopher Clark, Pradeep Dasigi, Nouha Dziri, Michal Guerquin, Hamish Ivison, Pang~Wei Koh, Jiacheng Liu, Saumya Malik, William Merrill, Lester James~V. Miranda, Jacob Morrison, Tyler Murray, Crystal Nam, Valentina Pyatkin, Aman Rangapur, Michael Schmitz, Sam Skjonsberg, David Wadden, Christopher Wilhelm, Michael Wilson, Luke Zettlemoyer, Ali Farhadi, Noah~A. Smith, and Hannaneh Hajishirzi. 2025.
\newblock \href {https://arxiv.org/abs/2501.00656} {{2 OLMo 2 Furious}}.
\newblock \emph{ArXiv preprint}, abs/2501.00656.

\bibitem[{Tian et~al.(2023)Tian, Mitchell, Zhou, Sharma, Rafailov, Yao, Finn, and Manning}]{tian2023just}
Katherine Tian, Eric Mitchell, Allan Zhou, Archit Sharma, Rafael Rafailov, Huaxiu Yao, Chelsea Finn, and Christopher Manning. 2023.
\newblock \href {https://doi.org/10.18653/v1/2023.emnlp-main.330} {{Just Ask for Calibration: Strategies for Eliciting Calibrated Confidence Scores from Language Models Fine-Tuned with Human Feedback}}.
\newblock In \emph{Proceedings of the 2023 Conference on Empirical Methods in Natural Language Processing}, pages 5433--5442, Singapore. Association for Computational Linguistics.

\bibitem[{van~der Waa et~al.(2020)van~der Waa, Schoonderwoerd, van Diggelen, and Neerincx}]{van2020interpretable}
Jasper van~der Waa, Tjeerd Schoonderwoerd, Jurriaan van Diggelen, and Mark Neerincx. 2020.
\newblock \href {https://doi.org/10.1016/j.ijhcs.2020.102493} {{Interpretable Confidence Measures for Decision Support Systems}}.
\newblock \emph{International Journal of Human-Computer Studies}, 144:102493.

\bibitem[{Wallsten et~al.(1993)Wallsten, Budescu, Zwick, and Kemp}]{wallsten1993preferences}
Thomas~S. Wallsten, David~V. Budescu, Rami Zwick, and Steven~M. Kemp. 1993.
\newblock \href {https://doi.org/10.3758/BF03334162} {{Preferences and Reasons for Communicating Probabilistic Information in Verbal or Numerical Terms}}.
\newblock \emph{Bulletin of the Psychonomic Society}, 31(2):135--138.

\bibitem[{Wang and Atanasova(2025)}]{wang-atanasova-2025-self}
Yingming Wang and Pepa Atanasova. 2025.
\newblock \href {https://doi.org/10.18653/v1/2025.emnlp-main.427} {{Self-Critique and Refinement for Faithful Natural Language Explanations}}.
\newblock In \emph{Proceedings of the 2025 Conference on Empirical Methods in Natural Language Processing}, pages 8481--8507, Suzhou, China. Association for Computational Linguistics.

\bibitem[{Wang et~al.(2024)Wang, Gangi~Reddy, Mujahid, Arora, Rubashevskii, Geng, Mohammed~Afzal, Pan, Borenstein, Pillai, Augenstein, Gurevych, and Nakov}]{wang-etal-2024-factcheck}
Yuxia Wang, Revanth Gangi~Reddy, Zain~Muhammad Mujahid, Arnav Arora, Aleksandr Rubashevskii, Jiahui Geng, Osama Mohammed~Afzal, Liangming Pan, Nadav Borenstein, Aditya Pillai, Isabelle Augenstein, Iryna Gurevych, and Preslav Nakov. 2024.
\newblock \href {https://doi.org/10.18653/v1/2024.findings-emnlp.830} {{Factcheck-Bench: Fine-Grained Evaluation Benchmark for Automatic Fact-checkers}}.
\newblock In \emph{Findings of the Association for Computational Linguistics: EMNLP 2024}, pages 14199--14230, Miami, Florida, USA. Association for Computational Linguistics.

\bibitem[{Warren et~al.(2025)Warren, Shklovski, and Augenstein}]{warren2025explainablefactchecking}
Greta Warren, Irina Shklovski, and Isabelle Augenstein. 2025.
\newblock \href {https://doi.org/10.48550/arXiv.2502.09083} {{Show Me the Work: Fact-Checkers' Requirements for Explainable Automated Fact-Checking}}.
\newblock In \emph{Proceedings of the CHI Conference on Human Factors in Computing Systems}, CHI '25, New York, NY, USA. Association for Computing Machinery.

\bibitem[{Warren et~al.(2026)Warren, Sun, Shklovski, and Augenstein}]{warren2026evidenceevaluatingroleevidence}
Greta Warren, Jingyi Sun, Irina Shklovski, and Isabelle Augenstein. 2026.
\newblock \href {https://arxiv.org/abs/2601.11387} {{Show Me the Evidence: Evaluating the Role of Evidence and Natural Language Explanations in AI-supported Fact-checking}}.
\newblock \emph{ArXiv preprint}, abs/2601.11387.

\bibitem[{Wei~Jie et~al.(2024)Wei~Jie, Satapathy, Goh, and Cambria}]{wei-jie-etal-2024-interpretable}
Yeo Wei~Jie, Ranjan Satapathy, Rick Goh, and Erik Cambria. 2024.
\newblock \href {https://aclanthology.org/2024.findings-naacl.138} {{How Interpretable are Reasoning Explanations from Prompting Large Language Models?}}
\newblock In \emph{Findings of the Association for Computational Linguistics: NAACL 2024}, pages 2148--2164, Mexico City, Mexico. Association for Computational Linguistics.

\bibitem[{Wiegreffe et~al.(2021)Wiegreffe, Marasovi{\'c}, and Smith}]{wiegreffe-etal-2021-measuring}
Sarah Wiegreffe, Ana Marasovi{\'c}, and Noah~A. Smith. 2021.
\newblock \href {https://doi.org/10.18653/v1/2021.emnlp-main.804} {{Measuring Association Between Labels and Free-Text Rationales}}.
\newblock In \emph{Proceedings of the 2021 Conference on Empirical Methods in Natural Language Processing}, pages 10266--10284, Online and Punta Cana, Dominican Republic. Association for Computational Linguistics.

\bibitem[{Windschitl and Wells(1996)}]{windschitl1996measuring}
Paul~D Windschitl and Gary~L Wells. 1996.
\newblock \href {https://doi.org/10.1037/1076-898X.2.4.343} {{Measuring Psychological Uncertainty: Verbal versus Numeric Methods.}}
\newblock \emph{Journal of Experimental Psychology: Applied}, 2(4):343.

\bibitem[{Xiong et~al.(2024)Xiong, Hu, Lu, Li, Fu, He, and Hooi}]{xiong2023can}
Miao Xiong, Zhiyuan Hu, Xinyang Lu, Yifei Li, Jie Fu, Junxian He, and Bryan Hooi. 2024.
\newblock \href {https://openreview.net/forum?id=gjeQKFxFpZ} {{Can LLMs Express Their Uncertainty? An Empirical Evaluation of Confidence Elicitation in LLMs}}.
\newblock In \emph{The Twelfth International Conference on Learning Representations, {ICLR} 2024, Vienna, Austria, May 7-11, 2024}. OpenReview.net.

\bibitem[{Yang et~al.(2025)Yang, Yoo, and Lee}]{yang-etal-2025-maqa}
Yongjin Yang, Haneul Yoo, and Hwaran Lee. 2025.
\newblock \href {https://aclanthology.org/2025.findings-naacl.325/} {{MAQA}: {Evaluating Uncertainty Quantification} in {LLM}s {Regarding Data Uncertainty}}.
\newblock In \emph{Findings of the Association for Computational Linguistics: NAACL 2025}, pages 5846--5863, Albuquerque, New Mexico. Association for Computational Linguistics.

\bibitem[{Yona et~al.(2024)Yona, Aharoni, and Geva}]{yona-etal-2024-large}
Gal Yona, Roee Aharoni, and Mor Geva. 2024.
\newblock \href {https://doi.org/10.18653/v1/2024.emnlp-main.443} {{Can Large Language Models Faithfully Express Their Intrinsic Uncertainty in Words?}}
\newblock In \emph{Proceedings of the 2024 Conference on Empirical Methods in Natural Language Processing}, pages 7752--7764, Miami, Florida, USA. Association for Computational Linguistics.

\bibitem[{Yuan et~al.(2025)Yuan, Sun, Zhang, F{\"a}rber, Eger, Atanasova, and Augenstein}]{yuan-etal-2025-graph}
Shuzhou Yuan, Jingyi Sun, Ran Zhang, Michael F{\"a}rber, Steffen Eger, Pepa Atanasova, and Isabelle Augenstein. 2025.
\newblock \href {https://doi.org/10.18653/v1/2025.emnlp-main.1494} {{Graph-Guided Textual Explanation Generation Framework}}.
\newblock In \emph{Proceedings of the 2025 Conference on Empirical Methods in Natural Language Processing}, pages 29362--29386, Suzhou, China. Association for Computational Linguistics.

\bibitem[{Zeng and Gao(2024)}]{zeng-gao-2024-justilm}
Fengzhu Zeng and Wei Gao. 2024.
\newblock \href {https://doi.org/10.1162/tacl_a_00649} {{J}usti{LM}: {Few-shot Justification Generation for Explainable Fact-Checking of Real-world Claims}}.
\newblock \emph{Transactions of the Association for Computational Linguistics}, 12:334--354.

\bibitem[{Zhang et~al.(2024{\natexlab{a}})Zhang, Liu, Basaldella, and Collier}]{zhang-etal-2024-luq}
Caiqi Zhang, Fangyu Liu, Marco Basaldella, and Nigel Collier. 2024{\natexlab{a}}.
\newblock \href {https://doi.org/10.18653/v1/2024.emnlp-main.299} {{LUQ}: {Long-text Uncertainty Quantification} for {LLM}s}.
\newblock In \emph{Proceedings of the 2024 Conference on Empirical Methods in Natural Language Processing}, pages 5244--5262, Miami, Florida, USA. Association for Computational Linguistics.

\bibitem[{Zhang et~al.(2024{\natexlab{b}})Zhang, Singh, Liu, Liu, Yu, Gao, and Zhao}]{zhang-etal-2024-attend}
Qingru Zhang, Chandan Singh, Liyuan Liu, Xiaodong Liu, Bin Yu, Jianfeng Gao, and Tuo Zhao. 2024{\natexlab{b}}.
\newblock \href {https://openreview.net/forum?id=xZDWO0oejD} {{Tell Your Model Where to Attend: Post-hoc Attention Steering for LLMs}}.
\newblock In \emph{The Twelfth International Conference on Learning Representations, {ICLR} 2024, Vienna, Austria, May 7-11, 2024}. OpenReview.net.

\bibitem[{Zhao et~al.(2024)Zhao, Wang, Wang, Cheng, Zhang, and Wong}]{zhao2024pacar}
Xiaoyan Zhao, Lingzhi Wang, Zhanghao Wang, Hong Cheng, Rui Zhang, and Kam-Fai Wong. 2024.
\newblock \href {https://aclanthology.org/2024.lrec-main.1099} {{PACAR}: {Automated Fact-Checking with Planning and Customized Action Reasoning Using Large Language Models}}.
\newblock In \emph{Proceedings of the 2024 Joint International Conference on Computational Linguistics, Language Resources and Evaluation (LREC-COLING 2024)}, pages 12564--12573, Torino, Italia. ELRA and ICCL.

\bibitem[{Zimmer(1983)}]{zimmer1983verbal}
Alf~C Zimmer. 1983.
\newblock \href {https://doi.org/10.1016/S0166-4115(08)62198-6} {{Verbal vs. Numerical Processing of Subjective Probabilities}}.
\newblock In \emph{Advances in psychology}, volume~16, pages 159--182. Elsevier.

\end{thebibliography}

\appendix
\section{Backbone model performance on public benchmarks}
\label{app:llm_perf}

Table~\ref{tab:llm_benchmarks} summarises the publicly reported
five-shot results on two standard reasoning benchmarks.  All figures are
taken verbatim from the official model cards or accompanying technical
reports. Figures are copied from the official model cards.

\begin{table*}[ht]
  \centering
  \begin{tabular}{lccc}
    \toprule
    \textbf{Model} & \textbf{Params} & \textbf{MMLU} & \textbf{GSM8K} \\
    \midrule
    Qwen2.5-14B-Instruct~\citep{qwen2.5} & 14.7 B & 79.7 & 90.2 \\
    Gemma-2-9B-IT~\citep{gemma_arxiv_2024}        & 9.0 B  & 71.3 & 68.6 \\
    OLMo-2-1124-13B-Instruct~\citep{olmo20242olmo2furious} & 13 B   & 67.5 & 54.2 \\
    \bottomrule
  \end{tabular}
    \caption{Benchmark scores on MMLU \citep{hendrycks2021measuring} and GSM8K \citep{cobbe2021training} are used to characterize instruction-following and reasoning strength.
}
  \label{tab:llm_benchmarks}
\end{table*}

These numbers corroborate our claim that Qwen2.5-14B-Instruct is the
strongest of the three for instruction-following and reasoning.

\section{Method: Selecting attention heads to steer}
\label{app:attention_heads_steering_selection}
Following~\citet{zhang-etal-2024-attend}, we steer only a selected subset of attention heads rather than all of them, because targeted steering yields larger gains in output quality.  Our selection criterion, however, differs from theirs: instead of ranking heads by their impact on task accuracy, we rank them by how strongly they affect the model’s \emph{predictive uncertainty} during fact-checking.

Concretely, for each fact-checking dataset chosen in this work(see details in \S\ref{sec:datasets}), $D$, we draw a validation subset $D_d$ with $|D_d| = 300$ examples.  For every input $X \in D_d$, we compute the model’s baseline uncertainty score $u(X)$ when it predicts the fact-checking label as stated in  \S\ref{sec:uncertainty_score_calculation}.  Then, for each attention head identified by layer $\ell$ and index $h$, we zero out that head, re-run the model, and measure the absolute change in uncertainty
$$
\Delta u(X, \ell, h) \;=\; \bigl|\,u(X)\;-\;u_{/o(l,h)}(X)\bigr|.
$$
Averaging $\Delta u(X, l, h)$ over all $X \in D_d$ yields a single importance score for head $(\ell,h)$.  We rank the heads by this score and keep the top $t$ heads for each dataset and each model. Note that we set $t = 100$ in line with the recommendation of~\citet{zhang-etal-2024-attend} and to balance steering effectiveness against the risk of degeneration.


\section{Conflict and Agreement Identification Details}
\subsection{Prompt Example for Assigning Relation Labels to Captured Span Interactions}
\label{app:relation_prompt}
To identify agreements and conflicts between the claim and the two evidence passages, we use the prompt in Figure \ref{fig:span-relation-prompt} to label each extracted span interaction (see \S\ref{sec:conflicts_extraction:method}).

\begin{figure}[t]
\centering
\begin{lstlisting}[
  basicstyle=\ttfamily\scriptsize,
  breaklines=true,
  frame=single,
  frameround=ffff
]
You are a helpful assistant. Your task:

1. Read the claim and its two evidence passages (E1, E2).
2. For each supplied span interaction, decide whether the two spans
   AGREE, DISAGREE, or are UNRELATED, taking the full context into account.
3. Output the span pairs exactly as given, followed by
   "relation: agree|disagree|unrelated".

Return format:
  1. "SPAN A" - "SPAN B"  relation: <agree|disagree|unrelated>
  2. ...
  3. ...

### SHOT 1  (annotated example)
Claim: [...]
Evidence 1: [...]
Evidence 2: [...]

Span interactions (to be labelled):
  1. "[...]" - "[...]"
  2. "[...]" - "[...]"
  3. "[...]" - "[...]"

Expected output:
  1. "[...]" - "[...]"  relation: ...
  2. "[...]" - "[...]"  relation: ...
  3. "[...]" - "[...]"  relation: ...

### SHOT 2   % omitted for brevity
### SHOT 3   % omitted for brevity

### NEW INSTANCE  (pre-filled for each new example)
Claim: {CLAIM}
Evidence 1: {E1}
Evidence 2: {E2}
Span interactions:
  1. "{SPAN1-A}" - "{SPAN1-B}"
  2. "{SPAN2-A}" - "{SPAN2-B}"
  3. "{SPAN3-A}" - "{SPAN3-B}"
\end{lstlisting}
\caption{Prompt template for span interaction relation labelling.}
\label{fig:span-relation-prompt}
\end{figure}

\subsection{Open-weight Alternative for Span-Interaction Relation Labeling}
\label{app:ablation_qwen_labeler}

This section studies how the choice of relation labeler affects downstream NLE quality in \ModelName{}.
While we use GPT-4o as the default relation labeler to reduce labeling noise, we show that a strong
open-weight alternative is also feasible, with only a small performance drop.

As discussed in Sec.~\ref{sec:conflicts_extraction:method}, \ModelName{} uses a relation labeler
$L_{\mathrm{rel}}$ to assign each extracted span interaction one of
$\{\textsc{agree}, \textsc{disagree}, \textsc{unrelated}\}$.
We considered two labeler options: (i) a closed-API model, which can reduce label noise
at the cost of API usage, and (ii) an open-weight model, which is fully local but may
yield slightly noisier labels and thus introduce additional noise in downstream NLE generation.

To quantify the impact of the labeler choice, we reran the pipeline with one open-weight alternative for
$L_{\mathrm{rel}}$: Qwen2.5-14B-Instruct, keeping all other components fixed.
Table~\ref{tab:labeler_ablation_qwen25} reports the results. 

\begin{table*}[ht]
\centering
\scalebox{0.75}{
\begin{tabular}{@{}l|cccc|cccc@{}}
\toprule
& \multicolumn{4}{c|}{\textbf{HealthVer}} & \multicolumn{4}{c}{\textbf{DRUID}} \\
\cmidrule(lr){2-5}\cmidrule(lr){6-9}
\textbf{Relation labeler}
& Faith. ($\uparrow$) & Span-Cov. ($\uparrow$) & Span-Ext. ($\downarrow$) & LEE ($\uparrow$)
& Faith. ($\uparrow$) & Span-Cov. ($\uparrow$) & Span-Ext. ($\downarrow$) & LEE ($\uparrow$) \\
\midrule
GPT-4o
& \textbf{0.033} & \textbf{0.44} & \textbf{0.53} & \textbf{0.80}
& \textbf{0.102} & \textbf{0.28} & \textbf{0.20} & \textbf{0.77} \\
Qwen2.5-14B-Instruct
& 0.025 & 0.41 & 0.58 & 0.78
& 0.076 & 0.25 & 0.25 & 0.75 \\
\bottomrule
\end{tabular}}
\caption{Effect of span-interaction \emph{relation labeler} choice. We rerun \ModelName\ while swapping only
$L_{\mathrm{rel}}$ (GPT-4o vs.\ Qwen2.5-14B-Instruct) and keeping all other components fixed
(backbone: Qwen2.5-14B-Instruct; variant: \modelvarianttwo{}; $k{=}3$).
An open-weight labeler is feasible with only a small drop in downstream NLE quality.
Bold indicates the better result per metric within each dataset.}

\label{tab:labeler_ablation_qwen25}
\end{table*}

Overall, Qwen2.5-14B-Instruct yields similar, though
slightly weaker, downstream NLE quality compared to GPT-4o across both datasets.
For instance, on HealthVer, Faithfulness drops from $0.033$ to $0.025$ and LEE from $0.80$ to $0.78$;
on DRUID, Faithfulness drops from $0.102$ to $0.076$ and LEE from $0.77$ to $0.75$.

\textbf{These results indicate that an open-weight labeler is feasible in \ModelName{}, and the framework is not
tied to a proprietary component.} In this work, we use GPT-4o in the main experiments because it provides
consistently stronger relation labeling and reduces the risk of error propagation from noisy relation labels
to the final explanations. We further discuss how upstream labeling errors can propagate and mitigation
strategies in App.~\ref{app:error_propagation}.

A natural concern is whether using a closed-API labeler compromises transparency.
We emphasize that \ModelName{} is \emph{white-box with respect to the backbone model}:
uncertainty scoring, span-interaction extraction, and attention steering require access to the backbone
model internals, which closed APIs do not provide.
Accordingly, running the \emph{entire} pipeline with GPT-4o would forgo these white-box signals
and reduce the method to surface-level prompting; prior work shows such verbalized uncertainty
can be unreliable, and fact-checking practice requires systems to ``show the work'' behind a decision
\citep{yona-etal-2024-large,tanneru2024uncertaintyLLMs,warren2025explainablefactchecking}.

GPT-4o is used only for the narrow task of pairwise relation judgment \emph{after} span interactions
have already been extracted from backbone internals; as such, it does not affect the transparency
of the core framework.





\subsection{Error Propagation Analysis}
\label{app:error_propagation}

We manually inspected 50 HealthVer instances (top-$3$ span pairs per instance; $N{=}150$ span pairs)
to assess (i) span extraction quality and (ii) relation labeling accuracy when using GPT-4o as $L_{\mathrm{rel}}$.

\paragraph{Span extraction quality.}
Most extracted spans are very short: 105/150 span pairs contain two-token spans (often salient nouns such as subjects/objects).
In 35/150 cases, at least one span is incomplete (e.g., ``D'' instead of ``vitamin D'') or semantically underspecified (e.g., ``is not''). Overall, the extractor tends to favor short, high-salience tokens, suggesting potential for future work to encourage more complete semantic units.

\paragraph{Relation labelling accuracy.}
On these 150 span pairs, GPT-4o achieves 86\% accuracy (129/150 correct).
Among the 21 errors, 12 are cases where an \textsc{agree} relation is mislabelled as \textsc{unrelated},
typically when one span is incomplete (e.g., ``D'' vs.\ ``vitamin D''), indicating that span completeness is a primary failure mode.

\paragraph{Propagation to downstream NLEs.}
Using Qwen2.5-14B-Instruct with CLUE-Span+Steering, 13/21 mislabeled interactions are explicitly mentioned in the generated NLEs,
showing that upstream errors can propagate into explanations.
Moreover, the subset of instances with mislabeled interactions exhibits lower faithfulness (mean Entropy-CCT 0.020)
than the overall HealthVer--Qwen CLUE-Span+Steering average (0.033; Table~1), suggesting that incorrect relation labels
can reduce explanation faithfulness.

This analysis suggests that fragmentary spans and relation-label noise can be an upstream error source and carry forward to the final NLEs, motivating mitigation strategies discussed in App.~\ref{app:error_propagation_mitigation} for future work.



\subsection{Mitigating Error Propagation}
\label{app:error_propagation_mitigation}

We outline several directions to mitigate error propagation in future work:
(1) \textbf{Improve span extraction:} add mild constraints during community detection so extracted spans more consistently capture complete semantic units rather than clipped tokens.
(2) \textbf{Improve relation labeling robustness:} use stronger prompting (broader local context, higher-quality example shots) or lightly fine-tune an open-weight model on a compact relation-judgment set.
(3) \textbf{Generation-time sanity checks:} add a lightweight consistency check that flags or down-weights span pairs judged internally inconsistent with the claim/evidence before producing the final NLE.

\section{Resources and Model Cards}
\label{app:resources}


Model cards: 
\href{https://huggingface.co/Qwen/Qwen2.5-14B-Instruct}{Qwen2.5-14B-Instruct (Hugging Face)};
\href{https://huggingface.co/google/gemma-2-9b-it}{Gemma-2-9B-IT (Hugging Face)};
\href{https://huggingface.co/allenai/OLMo-2-1124-13B-Instruct}{OLMo-2-1124-13B-Instruct (Hugging Face)};
\href{https://huggingface.co/deepseek-ai/DeepSeek-R1-Distill-Qwen-14B}{DeepSeek-R1-Distill-Qwen-14B (Hugging Face)};
\href{https://huggingface.co/MoritzLaurer/DeBERTa-v3-large-mnli-fever-anli-ling-wanli}{DeBERTa-v3 NLI (Hugging Face)};
\href{https://openai.com/index/hello-gpt-4o/}{GPT-4o (OpenAI)}.




\section{Computational Cost and Resource Requirements}
\label{app:compute_cost}

\paragraph{Setup.}
We report the computational overhead of \ModelName{} compared to \texttt{\baselinemodelname}.
Unless otherwise stated, measurements are collected using Qwen2.5-14B-Instruct on the HealthVer dataset
with a single NVIDIA A100 80GB GPU.
All values are reported \emph{per instance} (averaged over the evaluation set).

\subsection{Latency}
Table~\ref{tab:cost_latency} reports the average wall-clock time per instance for the three methods.
Both \texttt{\modelvariantone} and \texttt{\modelvarianttwo} are roughly 2 seconds slower per instance than
\texttt{\baselinemodelname}, and this overhead is entirely due to the additional span-interaction extraction and
relation labeling stage.

\begin{table*}[t]
\centering
\scalebox{0.85}{
\begin{tabular}{@{}l|cccc@{}}
\toprule
\textbf{Approach} & \textbf{Uncertainty Scoring (s)} & \textbf{Span Extraction + Labeling (s)} & \textbf{NLE Generation (s)} & \textbf{Total (s)} \\
\midrule
\texttt{\baselinemodelname} & 0.364 & --     & 14.8969 & 15.2609 \\
\texttt{\modelvariantone}   & 0.364 & 2.0452 & 14.7730 & 17.1822 \\
\texttt{\modelvarianttwo}   & 0.364 & 2.0452 & 14.7874 & 17.1966 \\
\bottomrule
\end{tabular}}
\caption{Average time cost per instance for the three NLE generation methods on HealthVer using Qwen2.5-14B-Instruct on a single NVIDIA A100 80GB GPU. ``--'' indicates that span extraction/labeling is not used by \texttt{\baselinemodelname}.}
\label{tab:cost_latency}
\end{table*}

\subsection{GPU memory usage}
Table~\ref{tab:cost_vram} reports peak GPU memory usage (VRAM).
During NLE generation, VRAM usage is similar across all methods.
\texttt{\modelvarianttwo} requires about 0.32 GiB more VRAM than \texttt{\baselinemodelname}, primarily because
the prompt is longer (it includes the extracted span interactions; see App.~\ref{app:span_prompt}).
Notably, \texttt{\modelvarianttwo} does not require more VRAM than \texttt{\modelvariantone}, since during generation
KV-cache and activations dominate memory and steering does not add a separate memory footprint beyond the base forward pass.

\begin{table*}[t]
\centering
\scalebox{0.85}{
\begin{tabular}{@{}l|ccc@{}}
\toprule
\textbf{Approach} & \textbf{Uncertainty Scoring (GiB)} & \textbf{Span Interaction Extraction (GiB)} & \textbf{NLE Generation (GiB)} \\
\midrule
\texttt{\baselinemodelname} & 29.71 & --    & 30.52 \\
\texttt{\modelvariantone}   & 29.71 & 29.72 & 30.84 \\
\texttt{\modelvarianttwo}   & 29.71 & 29.72 & 30.84 \\
\bottomrule
\end{tabular}}
\caption{Average peak VRAM per instance for the three NLE generation methods (HealthVer, Qwen2.5-14B-Instruct, NVIDIA A100 80GB).}
\label{tab:cost_vram}
\end{table*}

\subsection{Memory requirement for steering 100 heads}
For selecting the most important 100 attention heads (App.~\ref{app:attention_heads_steering_selection}), the peak VRAM is 29.8 GiB.
For steering these heads during NLE generation (\texttt{\modelvarianttwo}), the average peak VRAM per instance is 30.84 GiB.
Thus, a GPU with 40GB of memory satisfies the requirement for attention steering across 100 heads.

\subsection{GPT-4o API cost for relation labeling}
When using GPT-4o as $L_{\text{rel}}$, the total relation-labeling cost scales linearly with the number of labeled instances. In our main experiments, the cumulative cost across datasets and backbones was approximately \$30 (about \$0.003 per instance),
with an average input length of $\sim$1k tokens and an average output length of $\sim$50 tokens per call.

\section{Methodology Supplement: Faithfulness Evaluation Details}
\label{app:method_supp}

This supplement provides implementation and statistical details for the Entropy-CCT faithfulness
metric introduced in \S\ref{sec:faithfulness:method}. We describe (i) how counterfactual perturbations
are generated, (ii) how Entropy-CCT relates to the original CCT, and (iii) how we test statistical
significance and report results.

\subsection{Entropy-CCT Perturbation Generation}
\label{app:perturbation_details}
To evaluate how faithfully each NLE reflects model uncertainty, we generate multiple counterfactuals per instance, following \citet{atanasova-etal-2020-generating-fact} and \citet{siegel-etal-2024-probabilities} (see \S\ref{sec:faithfulness:method}).  For every input, comprising one claim and two evidence passages, we first tag part‐of‐speech with
spaCy~\citep{honnibal_montani_spacy2_2017} (\texttt{en\_core\_web\_sm}), then choose four random insertion sites (uniformly among tokens tagged as NOUN or VERB).  At each site we insert either (i) a random adjective before a noun or (ii) a random adverb before a verb.  The candidate modifiers are drawn uniformly from WordNet~\citep{miller_1995_wordnet} using NLTK~\citep{bird_klein_loper_2009_nltk}: we construct adjective/adverb candidate lists by collecting all lemma names from all WordNet adjective/adverb synsets, respectively.
Because we sample three random candidates for each of the four positions, this procedure yields \(4 \times 3 = 12\) perturbations per instance, providing a sufficient set for the subsequent Entropy-CCT evaluation, in which we check whether the NLE mentions the inserted word and compute the correlation between that mention and the uncertainty change induced by each perturbation.

\subsection{Entropy‑CCT vs. Original CCT}
\label{app:difference_from_cct}
CCT~\cite{siegel-etal-2024-probabilities} evaluates faithfulness by perturbing the input and measuring
the resulting change in the model's predicted label distribution. Specifically, it uses the Total Variation Distance (TVD) between the pre- and post-perturbation distributions $P(\cdot\mid X)$ and $P(\cdot\mid X')$:
\begin{equation}
\mathrm{TVD}(P,Q)=\frac{1}{2}\sum_i |P_i - Q_i|.
\end{equation}

Entropy-CCT is a minimal adaptation of CCT that preserves the same perturbation procedure and
the same correlational test: we correlate the perturbation impact with whether the inserted token is
mentioned in the NLE. The only change is the signal we correlate. Instead of TVD (probability-shift),
we use Absolute Entropy Change (AEC), i.e., the change in predictive entropy, to directly target
uncertainty (our object of explanation). 




\subsection{Statistical Testing \& Significance}
\label{app:faithfulness_stats}

We test statistical significance of $r_{\text{pb}}$ using the standard Pearson correlation $t$-test.
Each perturbation constitutes one observation: for each dataset
we evaluate 600 instances with 12 perturbations per instance, yielding $n=600\times 12=7{,}200$
observations (App.~\ref{app:perturbation_details}). Because the point-biserial correlation is
algebraically equivalent to Pearson correlation between a continuous and a binary variable, we use
the standard Pearson $t$-test for $H_0: r_{\text{pb}}=0$.



\textbf{Interpreting $r_{\text{pb}}$ and $\Delta r_{\text{pb}}$}. The Entropy-CCT score is the point-biserial correlation \citep{tate1954correlation} between the absolute entropy change $|\Delta u|$ and the binary mention flag $m$. Because it is mathematically identical to a Pearson $r$ computed between one continuous and one binary variable, it obeys $-1\le r_{\text{pb}}\le1$. When $r_{\text{pb}}=0$, it means the high- and low-impact perturbations are mentioned equally often.  If the two strata are roughly balanced, every $+0.01$ in $r_{\text{pb}}$ increases the probability that a truly uncertainty-influential token is mentioned by about one percentage point (pp).  A \emph{gain} $\Delta r_{\text{pb}}$ therefore translates to an \emph{absolute} improvement of $\approx |\Delta r_{\text{pb}}| \times 100$~pp in mention rate. For instance, moving from $-0.08$ to $+0.06$ is a swing of $0.14$, corresponding to about $14$~pp.

\textbf{Significance testing}.
Because the point-biserial is a Pearson correlation, the familiar $t$--test applies:

\begin{align}
  t &= r_{\text{pb}}\,
       \sqrt{\frac{n-2}{1-r_{\text{pb}}^{2}}}, \\[4pt]
  t &\sim t_{(n-2)}
       \qquad\text{under } H_{0}\!: r_{\text{pb}} = 0.
\end{align}

With $n=7,200$ we have $\text{df}=7,198$; the critical two-sided
values are $|t|>1.96$ for $p<0.05$ and $|t|>2.58$ for $p<0.01$.


\textbf{Faithfulness with significance results}. Table ~\ref{tab:faith_sig_7200} shows the point-biserial
coefficients $r_{\text{pb}}$, which is our faithfulness measurement for model uncertainty(See, Eq.\ref{eq:entropy-cct}), the associated $t$ statistics, and two-sided $p$ values for every model--method pair.  Values that meet the stricter
$p<0.01$ criterion are highlighted in bold.

\begin{table*}[t]          
\centering
\footnotesize
\setlength{\tabcolsep}{6pt} 
\begin{tabular}{@{}l l r r r@{}}
\toprule
\textbf{Model} & \textbf{Method} & $r_{\text{pb}}$ & $t$ & $p$ \\
\midrule
\multicolumn{5}{c}{\textbf{HealthVer}}\\
\midrule
Qwen2.5-14B-Instruct & \textbf{\texttt{\baselinemodelname}}      & $-0.028$ & $-2.38$ & $1.7\times10^{-2}$ \\
            & \textbf{\texttt{\modelvariantone}}        & $+0.006$ & $+0.51$ & $6.1\times10^{-1}$ \\
            & \textbf{\texttt{\modelvarianttwo}}        & $+0.033$ & $+2.80$ & $\mathbf{5.1\times10^{-3}}$ \\[2pt]
OLMo-2-1124-13B-Instruct  & \textbf{\texttt{\baselinemodelname}}      & $-0.100$ & $-8.53$ & $\mathbf{<10^{-15}}$ \\
            & \textbf{\texttt{\modelvariantone}}        & $+0.005$ & $+0.42$ & $6.7\times10^{-1}$ \\
            & \textbf{\texttt{\modelvarianttwo}}        & $+0.020$ & $+1.70$ & $9.0\times10^{-2}$ \\[2pt]
Gemma-2-9B-IT  & \textbf{\texttt{\baselinemodelname}}      & $-0.105$ & $-8.96$ & $\mathbf{<10^{-15}}$ \\
            & \textbf{\texttt{\modelvariantone}}        & $+0.007$ & $+0.59$ & $5.5\times10^{-1}$ \\
            & \textbf{\texttt{\modelvarianttwo}}        & $+0.021$ & $+1.78$ & $7.5\times10^{-2}$ \\
\midrule
\multicolumn{5}{c}{\textbf{DRUID}}\\
\midrule
Qwen2.5-14B-Instruct & \textbf{\texttt{\baselinemodelname}}      & $-0.080$ & $-6.81$ & $\mathbf{9.8\times10^{-12}}$ \\
            & \textbf{\texttt{\modelvariantone}}        & $+0.089$ & $+7.58$ & $\mathbf{3.4\times10^{-14}}$ \\
            & \textbf{\texttt{\modelvarianttwo}}        & $+0.102$ & $+8.70$ & $\mathbf{<10^{-15}}$ \\[2pt]
OLMo-2-1124-13B-Instruct  & \textbf{\texttt{\baselinemodelname}}      & $-0.130$ & $-11.12$ & $\mathbf{<10^{-15}}$ \\
            & \textbf{\texttt{\modelvariantone}}        & $+0.014$ & $+1.19$ & $2.3\times10^{-1}$ \\
            & \textbf{\texttt{\modelvarianttwo}}        & $+0.099$ & $+8.44$ & $\mathbf{<10^{-15}}$ \\[2pt]
Gemma-2-9B-IT  & \textbf{\texttt{\baselinemodelname}}      & $-0.120$ & $-10.26$ & $\mathbf{<10^{-15}}$ \\
            & \textbf{\texttt{\modelvariantone}}        & $+0.043$ & $+3.65$ & $\mathbf{2.6\times10^{-4}}$ \\
            & \textbf{\texttt{\modelvarianttwo}}        & $+0.098$ & $+8.35$ & $\mathbf{<10^{-15}}$ \\
\bottomrule
\end{tabular}
\caption{Detailed faithfulness evaluation results for baseline method \textbf{\texttt{\baselinemodelname}}, and two variants of our \modelname\ framework \textbf{\texttt{\modelvariantone}} and \textbf{\texttt{\modelvarianttwo}} on Healthver and Druid dataset based on Qwen2.5-14B-Instruct(\citet{qwen2.5}), OLMo-2-1124-13B-Instruct(\citet{olmo20242olmo2furious})and Gemma-2-9B-IT(\citet{gemma_arxiv_2024}).
Point-biserial correlation $r_{\text{pb}}$ is our Entropy-CCT measurement(\S\ref{sec:faithfulness:method}), along with $t$ statistic and two-sided $p$-value for each model--method pair ($n=7{,}200$, $df=7{,}198$). Entries with $p<0.01$ are bold.}
\label{tab:faith_sig_7200}
\end{table*}

Across both datasets and all three backbones, the \textbf{\texttt{\baselinemodelname}} exhibits negative correlations, implying a \emph{non-faithful} tendency to highlight low-impact tokens in the generated NLEs, with $\text{mean}= -0.094$.  The prompt-only variant of our \modelname\ framework \textbf{\texttt{\modelvariantone}} neutralises this bias and turns the average into $+0.027$; two of its coefficients are clear $p<0.01$, indicating a modest but significant improvement regarding faithfulness.

The full \textbf{\texttt{\modelvarianttwo}} variant pushes the mean to
$+0.062$ and achieves $p<0.01$ in four of six settings.  Interpreting
these numbers using the discussion above, the
switch from $-0.094$ to $+0.062$ yields an \emph{absolute} increase of $(0.062-(-0.094))\times100!\approx16$, pp in the probability that a truly uncertainty-influential token is named in the NLE, which is easily noticeable in
qualitative inspection.

The consistently positive, statistically significant gains therefore
substantiate the claim made in the main text: \modelname\ produces
markedly more faithful NLEs towards model uncertainty than the
\textbf{\texttt{\baselinemodelname}}, and the steer variant is particularly
beneficial for models that initially struggle with uncertainty
attribution.

\section{Prompt template for \textbf{\texttt{\baselinemodelname}}, \textbf{\texttt{\modelvariantone}} and \textbf{\texttt{\modelvarianttwo}} on Healthver and Druid dataset}
\label{app:explanation_prompt}
We designed two prompt templates for our experiments. The baseline prompt (Figure  \ref{fig:baseline-prompt}) gives the model no span interactions; instead, it must first identify the relevant agreements or conflicts and then discuss them in its explanation.
In contrast, the prompt used by our \modelname\ framework (Figure \ref{fig:span-prompt}) supplies the three pre-extracted span interactions (\S\ref{sec:conflicts_extraction:method}). The model is explicitly instructed to base its explanation on these spans, ensuring that the rationale remains grounded in the provided evidence.

\subsection{Prompt template for \textbf{\texttt{\baselinemodelname}}}
\label{app:baseline_prompt}
To generate NLEs about model uncertainty without span-interaction guidance, we craft a three-shot prompt that instructs the model to identify the interactions most likely to affect its uncertainty and to explain how these relations they represent affect it. (See Figure \ref{fig:baseline-prompt}).

\begin{figure}[t]  
\centering
\begin{lstlisting}[
  basicstyle=\ttfamily\scriptsize,
  breaklines=true,
  frame=single,
  frameround=ffff
]
You are a helpful assistant. Your tasks:
1. Determine the relationship between the claim and the two evidence passages.
2. Explain your prediction's uncertainty by identifying the three most
   influential span interactions from Claim-Evidence 1, Claim-Evidence 2,
   and Evidence 1-Evidence 2, and describing how each interaction's relation
   (agree, disagree, or unrelated) affects your overall confidence.
Return format: [Prediction] [Explanation]

### SHOT 1
Input
  Claim: [...]
  Evidence 1: [...]
  Evidence 2: [...]
Output
  [Prediction: ...] [Explanation: ...]

### SHOT 2   % omitted for brevity
### SHOT 3   % omitted for brevity

### NEW INSTANCE
Claim: {CLAIM}
Evidence 1: {E1}
Evidence 2: {E2}
Your answer:
\end{lstlisting}
\caption{Three-shot prompt for \textbf{\texttt{\baselinemodelname}} (Shots 2-3 omitted) on the HealthVer and DRUID datasets.}
\label{fig:baseline-prompt}
\end{figure}

\subsection{Prompt template for \textbf{\texttt{\modelvariantone}} and \textbf{\texttt{\modelvarianttwo}}}
\label{app:span_prompt}
To generate NLEs about model uncertainty with the span-interaction guidance, we craft a three-shot prompt that instructs the model to discuss how these interactions, along with the relations they represent, affect its uncertainty. (See Figure \ref{fig:span-prompt}).

\begin{figure}[t]  
\centering
\begin{lstlisting}[
  basicstyle=\ttfamily\scriptsize,
  breaklines=true,
  frame=single,
  frameround=ffff
]
You are a helpful assistant. Your tasks:
1. Determine the relationship between the claim and the two evidence passages.
2. Explain your prediction's uncertainty by referring to the three span
   interactions provided below (Claim-Evidence 1, Claim-Evidence 2,
   Evidence 1-Evidence 2) and describing how each interaction's relation
   (agree, disagree, or unrelated) affects your overall confidence.
Return format: [Prediction] [Explanation]

### SHOT 1
Input:
  Claim: [...]
  Evidence 1: [...]
  Evidence 2: [...]
  Span interactions:
    1. ''[...]'' - ''[...]''  (C-E1)  relation: [...]
    2. ''[...]'' - ''[...]''  (C-E2)  relation: [...]
    3. ''[...]'' - ''[...]''  (E1-E2) relation: [...]
Output:
  [Prediction: ...] [Explanation: ...]

### SHOT 2   % omitted for brevity
### SHOT 3   % omitted for brevity

### NEW INSTANCE
Claim: {CLAIM}
Evidence 1: {E1}
Evidence 2: {E2}
Span interactions (pre-filled):
    1. ''{SPAN1-A}'' - ''{SPAN1-B}''  (C-E1)  relation: {REL1}
    2. ''{SPAN2-A}'' - ''{SPAN2-B}''  (C-E2)  relation: {REL2}
    3. ''{SPAN3-A}'' - ''{SPAN3-B}''  (E1-E2) relation: {REL3}
Your answer:
\end{lstlisting}
\caption{Three-shot prompt for \textbf{\texttt{\modelvariantone}} and \textbf{\texttt{\modelvarianttwo}} (Shots 2-3 omitted) on the HealthVer and DRUID datasets.}
\label{fig:span-prompt}
\end{figure}

\section{Extended Evaluation Results}
\label{app:extended_evaluation}

\subsection{Extended Evaluation Results for Scenarios Involving Three Evidence pieces}
\label{app:3_evidence}

\begin{table}[t]
\centering
\scalebox{0.89}{
\begin{tabular}{lccc}
\toprule
\textbf{Metrics} 
& \textbf{\texttt{\baselinemodelnameshortened}}
& \textbf{\texttt{\modelvariantoneshortened}}
& \textbf{\texttt{\modelvarianttwoshortened}} \\
\midrule
Faith.\ ($\uparrow$)      & 0.08 & 0.10 & \textbf{0.13} \\
Span-Cov.\ ($\uparrow$)   & --   & 0.25 & \textbf{0.41} \\
Span-Ext.\ ($\downarrow$) & --   & \textbf{0.23} & 0.27 \\
LEE ($\uparrow$)          & 0.80 & \textbf{0.86} & \textbf{0.86} \\
\bottomrule
\end{tabular}}
\caption{Automatic evaluation for three types of NLE on HealthVer using \textbf{Qwen2.5-14B-Instruct} when one claim and three pieces of evidence are involved: \textbf{\texttt{\baselinemodelnameshortened}}(\textbf{\texttt{\baselinemodelname}}) is the baseline method, \textbf{\texttt{\modelvariantoneshortened}} (\textbf{\texttt{\modelvariantone}}) and \textbf{\texttt{\modelvarianttwoshortened}}(\textbf{\texttt{\modelvarianttwo}}) are the two variants of our \modelname\ framework.}
\label{tab:3_evidences}
\end{table}

To verify the generalizability of our framework in scenarios involving multiple evidence pieces, we conduct an experiment on the Healthver dataset using the Qwen2.5‑14B‑Instruct model, pairing each claim with three pieces of evidence instead of two (as shown in Tab. \ref{tab:3_evidences}). Note that we select the top‑$k=6$ span interactions for NLE generation, taking the top‑1 interaction from each pairwise combination among the $C$, $E_1$, $E_2$, and $E_3$ input parts.

Overall, we observe a similar trend with three pieces of evidence as with two (see \S\ref{sec:automatic_results}): \textbf{\modelvarianttwo} and \textbf{\modelvariantone} show better Faithfulness and Label-Explanation Entailment than \textbf{\baselinemodelname}. Interestingly, the Faithfulness of all three explanation types increases relative to the two‑evidence setting, from -0.028 to 0.08 for \textbf{\baselinemodelname}, from 0.006 to 0.10 for \textbf{\modelvariantone}, and from 0.033 to 0.13 for \textbf{\modelvarianttwo}. This may be because, when NLEs are grounded in richer evidence, they capture more nuances in both claim-evidence and inter‑evidence interactions, thereby more faithfully reflecting the model’s fact‑checking decision.

It is notable that \textbf{\modelvarianttwo} includes more targeted span interactions (higher Span‑Coverage: 0.41 > 0.27) as well as more redundant span interactions (higher Span‑Extraneous: 0.27 > 0.23) than \textbf{\modelvariantone}. This suggests that, although \textbf{\modelvarianttwo} better focuses on target interactions, it still suffers from redundancy due to the inclusion of more irrelevant interactions, which we call for future work to address.

\subsection{Extended Evaluation Results on Reasoning Model}
\label{app:reasoning_model}

To demonstrate that our framework generalizes to reasoning models trained primarily via large‑scale reinforcement learning, and designed to encourage chain‑of‑thought exploration, we adopt \textit{DeepSeek‑R1‑Distill‑Qwen‑14B}\footnote{\url{https://huggingface.co/deepseek-ai/DeepSeek-R1-Distill-Qwen-14B}} \citep{deepseek_r1_distill_qwen_14b} as the base model and evaluate it on the HealthVer dataset in the setting with one claim and two pieces of evidence (see Tab.~\ref{tab:reasoning_model}).

\begin{table}[t]
\centering
\scalebox{0.9}{
\begin{tabular}{lccc}
\toprule
\textbf{Metrics} 
& \textbf{\texttt{\baselinemodelnameshortened}}
& \textbf{\texttt{\modelvariantoneshortened}}
& \textbf{\texttt{\modelvarianttwoshortened}} \\
\midrule
Faith.\ ($\uparrow$)      & 0.12 & 0.16 & \textbf{0.18} \\
Span-Cov.\ ($\uparrow$)   & --   & 0.58 & \textbf{0.67}\\
Span-Ext.\ ($\downarrow$) & --   & \textbf{0.59} & 0.64 \\
LEE ($\uparrow$)          & 0.88 & \textbf{0.90} & 0.89 \\
\bottomrule
\end{tabular}}
\caption{Automatic evaluation for three types of NLE on \textsc{HealthVer} using \textbf{DeepSeek-R1-Distill-Qwen-14B} when one claim and two pieces of evidence are involved: \textbf{\texttt{\baselinemodelnameshortened}}(\textbf{\texttt{\baselinemodelname}}) is the baseline method, \textbf{\texttt{\modelvariantoneshortened}} (\textbf{\texttt{\modelvariantone}}) and \textbf{\texttt{\modelvarianttwoshortened}}(\textbf{\texttt{\modelvarianttwo}}) are the two variants of our \modelname\ framework.}
\label{tab:reasoning_model}
\end{table}

Overall, both variants of our CLUE model, \textbf{\texttt{\modelvariantone}} and \textbf{\texttt{\modelvarianttwo}}, show higher Faithfulness and Label‑Explanation Entailment than the baseline method \textbf{\texttt{\baselinemodelname}}, confirming the effectiveness of our framework. It is also notable that, compared with the evaluation results on other models in Tab.~\ref{tab:healthver_druid_results}, all three types of explanations generated with the DeepSeek‑R1‑Distill‑Qwen‑14B model achieve the best performance across metrics, especially the highest Faithfulness and Span‑Coverage. The former implies that reinforcement-learning-trained reasoning models may be more capable of generating highly faithful explanations with the thinking step; the latter suggests that this type of model follows our interaction-guided instructions more effectively. Consequently, both \textbf{\texttt{\modelvariantone}} and \textbf{\texttt{\modelvarianttwo}} include more target-span interactions, yielding higher Span-Coverage.

Comparing the two variants, \textbf{\texttt{\modelvarianttwo}} exhibits higher Span‑Coverage (0.67 > 0.58) than \textbf{\texttt{\modelvariantone}}, but also more Span‑Extraneous (0.64 > 0.59), suggesting that \textbf{\texttt{\modelvarianttwo}} tends to include redundant interactions while focusing on target interactions. \textbf{\texttt{\modelvarianttwo}} also achieves the highest Faithfulness (0.18 > 0.16), consistent with its higher Span‑Coverage, but yields similar Label‑Explanation Entailment to \textbf{\texttt{\modelvariantone}} (0.89 vs.\ 0.90), likely due to the increased Span‑Extraneous. Addressing this trade‑off is left for future work.

\section{Human Evaluation Details}
\subsection{Participants and Materials}
\label{app:participants_and_materials}
\paragraph{Participants}
We recruited N=12 participants from \href{https://www.prolific.com/}{Prolific}, screened to be native English speakers from Australia, Canada, Ireland, New Zealand, the United Kingdom, and the United States. The study was approved by our institution’s
Research Ethics Committee (reference number anonymised).

\begin{figure}[b]
    \centering
    \includegraphics[width=\linewidth]{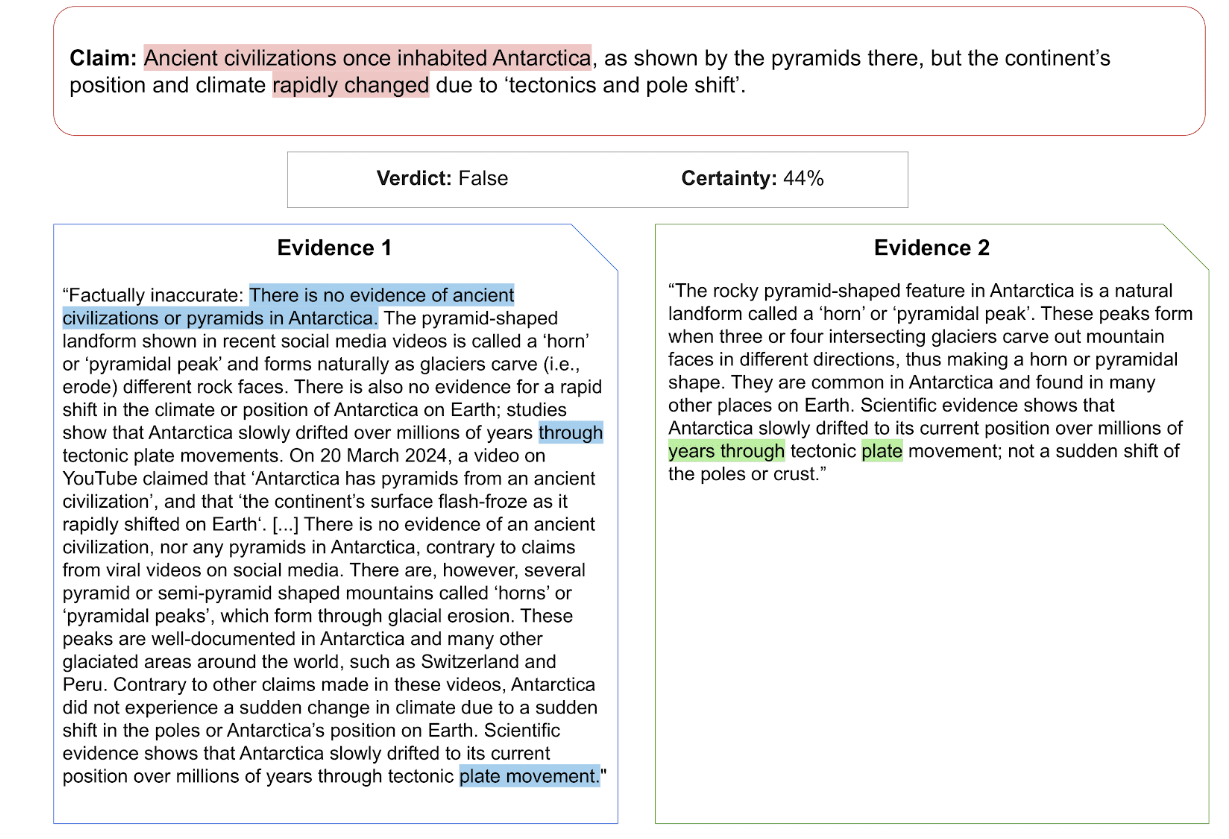}
    \includegraphics[width=\linewidth]{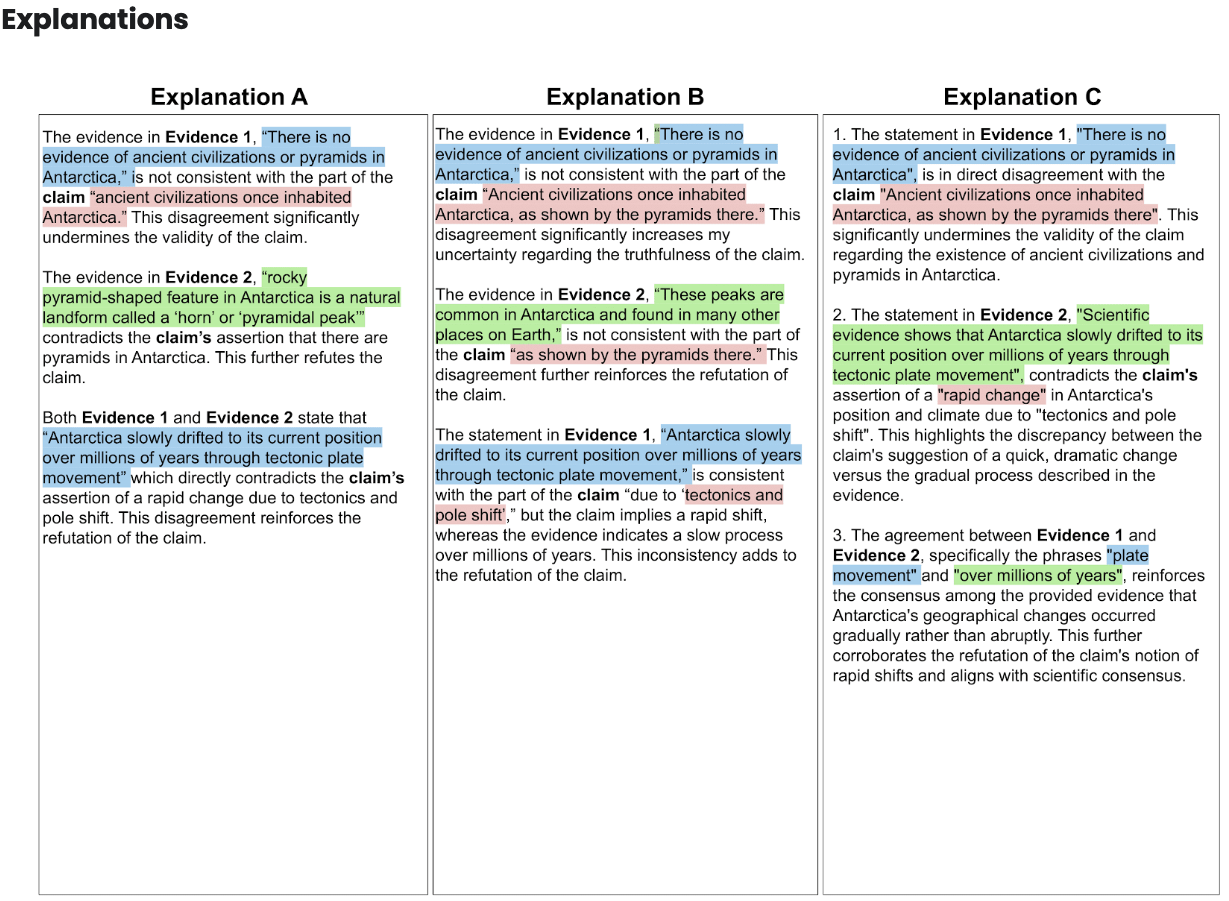}
    \caption{Example of human evaluation set-up. Explanation A was generated using \baselinemodelname, Explanation B by \modelvariantone, and Explanation C by \modelvarianttwo}
    \label{fig:human_eval}
\end{figure}

\paragraph{Materials} 
Forty claims (20 from DRUID, 20 from HealthVer) were selected at random for evaluation, while 120 explanations were evaluated in total.
For each instance, participants were provided with a claim, two evidence documents, model verdict, model numerical certainty, and three alternative explanations (see Figure \ref{fig:human_eval} in \ref{app:example_annotation}).
The explanations presented to participants were those generated using Qwen2.5-14b-instruct \cite{qwen2.5} based on its automatic evaluation performance.

Each participant evaluated explanations for 10 instances (5 labelled `True', 5 labelled `False'), in addition to two attention check instances which were used to screen responses for quality.
To mitigate the difficulty and cognitive demands of the task, we kept the number of instances per participant low, and removed participants who failed attention checks from the data.
In order to minimise the risk of bias,
each explanation presented to participants was assigned a neutral label: ‘Explanation A’, ‘Explanation B’, ‘Explanation C’, so that participants could not infer anything about how the explanations were generated or indicate that one should be favoured over another.
Each explanation was also presented in a different, random order for each instance (e.g., ABC for Claim 1, CAB for Claim 2, etc.) to mitigate order effects such as primacy bias or recency bias. The order in which the response options ‘Explanation A’, ‘Explanation B’, ‘Explanation C’ also appeared in the same random order.

\paragraph{Procedure} Participants read information about the study (see \ref{app:instructionseval}) and provided informed consent (see \ref{app:consentform}) before reading detailed task instructions and completing a practice example of the task (see \ref{app:instructs}). 
Participants then progressed through the study at their own pace.
The task took approximately 20 minutes, and participants were paid £3 for their work.

\subsection{Human Evaluation Results} \label{app:eval_results}
\subsubsection{Interrater agreement}
\label{app:IAA}

In line with similar NLE evaluations carried out by previous studies (e.g., \cite{atanasova-etal-2020-generating-fact}), interrater agreement (Kendall's W \cite{kendall1939problem}) was moderate to low (see Table \ref{tab:IAA}). We attribute this to the relative complexity of the task and individual differences in how the information was perceived.

\begin{table}[]
\scalebox{0.9}{
\begin{tabular}{lrrrr}
\hline
                & \multicolumn{2}{c}{\textbf{DRUID}}                    & \multicolumn{2}{c}{\textbf{HealthVer}}                \\
\textbf{}       & \multicolumn{1}{c}{Set A} & \multicolumn{1}{c}{Set B} & \multicolumn{1}{c}{Set A} & \multicolumn{1}{c}{Set B} \\ \hline
Helpfulness     & .016                     & .079                     & .003                     & .013                     \\
Consistency     & .44                      & .058                     & .017                    & .016                     \\
Non-redundancy  & .005                     & .084                     & .005                     & .019                     \\
Coverage        & .494                     & .113                     & .018                     & .027                     \\
Overall Quality & .005                     & .158                     & .01                      & .002                     \\ \hline
\end{tabular}
}
\caption{Interrater agreement (Kendall's W) for human evaluation}
\label{tab:IAA}
\end{table}

\subsubsection{CLUE Variant preferences}
\label{app:CLUE_variants}

\begin{table}[t]
\centering
\scalebox{0.9}{
\begin{tabular}{lccc}
\toprule
& \textbf{\texttt{\baselinemodelnameshortened}} 
& \textbf{\texttt{\modelvariantoneshortened}} 
& \textbf{\texttt{\modelvarianttwoshortened}} \\
\midrule
\multicolumn{4}{l}{\textbf{Helpfulness}} \\
Overall   & 2.025                                 & 1.892                             & \textbf{1.867}                      \\
DRUID     & 1.9                                   & 1.917                             & \textbf{1.767}                      \\
HealthVer & 2.15                                  & \textbf{1.867}                    & 1.967                               \\
\modelblocksep
\multicolumn{4}{l}{\textbf{Consistency}} \\
Overall   & 1.875                                 & \textbf{1.783}                    & 1.817                               \\
DRUID     & 1.717                                 & 1.75                              & \textbf{1.617}                      \\
HealthVer & 2.033                                 & \textbf{1.817}                    & 2.017                               \\ 
\addlinespace
\multicolumn{4}{l}{\textbf{Non-redundancy}} \\
Overall   & 2.05                                  & 1.908                             & \textbf{1.833}                      \\
DRUID     & 1.983                                 & 1.983                             & \textbf{1.683}                      \\
HealthVer & 2.117                                 & \textbf{1.833}                    & 1.983                               \\
\addlinespace
\multicolumn{4}{l}{\textbf{Coverage}} \\
Overall   & 1.967                                 & 1.775                             & \textbf{1.758}                      \\
DRUID     & 1.767                                 & 1.75                              & \textbf{1.617}                      \\
HealthVer & 2.167                                 & \textbf{1.8}                      & 1.9                                 \\
\addlinespace
\multicolumn{4}{l}{\textbf{Overall Quality}} \\
Overall   & 1.967                                 & \textbf{1.908}                    & 1.925                               \\
DRUID     & 1.9                                   & 1.9                               & \textbf{1.817}                      \\
HealthVer & 2.033                                 & \textbf{1.917}                    & 2.033                               \\
\bottomrule
\end{tabular}}
\caption{Mean Average Rank (MAR) for the five human-evaluation criteria applied to explanations from \textbf{Qwen2.5-14B-Instruct} on the \textsc{HealthVer} and \textsc{DRUID} datasets (chosen for its high faithfulness; see §\ref{sec:automatic_results}).  
\textbf{\texttt{\baselinemodelname}}, \textbf{\texttt{\modelvariantone}} (\textbf{\texttt{\modelvariantoneshortened}}), and \textbf{\texttt{\modelvarianttwo}} (\textbf{\texttt{\modelvarianttwoshortened}}) are compared.  
Lower MAR means a better (higher) average rank; the best score in each row is boldfaced.}
\label{tab:human_eval_ranks}
\end{table}

Table \ref{tab:human_eval_ranks} shows the mean rank assigned by participants to each explanation variant.
The explanations generated by \modelname\ were preferred by our participants to those generated using \textbf{\texttt{\baselinemodelname}}: the \textbf{explanations generated by \textbf{\texttt{\modelvarianttwo}} were rated as most helpful, highest coverage, and containing the least amount of redundant information}, while \textbf{those from \texttt{\modelvariantone} were judged to have the highest consistency and overall quality}.

Finally, we observed slight variation between datasets: \textbf{\texttt{\modelvarianttwo}} tended to be rated higher than \textbf{\texttt{\modelvariantone}} for DRUID, and vice versa for HealthVer.
This may arise from differences in length and complexity of the input: DRUID evidence documents, retrieved from heterogeneous online sources and often consisting of longer form new articles, may have benefited from attention steering more than HealthVer evidence documents which consist of focused, shorter extracts from scientific abstracts.

\subsection{Human Evaluation Information Screen}
\label{app:instructionseval}

Thank you for volunteering to participate in this study! Before you decide whether you wish to take part, please read this information screen carefully. 

\noindent \textbf{1. What is the project about?}

\noindent Our goal is to make sure that AI fact-checking systems can explain the decisions they produce in ways that are understandable and useful to people. This survey is part of a project to help us understand 
[anonymised]

\noindent \textbf{2. What does participation entail?}

\noindent You are invited to help us explore what kinds of explanations work better in fact-checking. In this task you will see claims, an AI system’s prediction about whether this claim is true or false and corresponding evidence used to make the prediction. You will also see an explanation for why the AI system is certain or uncertain about its prediction to help you decide how to interpret the true/false prediction. We ask you to evaluate the explanations along 5 different dimensions (the detailed explanation of the task is on the next page).
All participants who complete the survey will receive a payment of £3. There is no cost to you for participating. You may refuse to participate or discontinue your involvement at any time without penalty. 

\noindent \textbf{3. Source of funding} \\
\noindent This project has received funding from [anonymised]

\noindent \textbf{4. Consenting to participate in the project and withdrawing from the research} \\
\noindent You can consent to participating in this study by ticking the box on the next page of the study.  Participation in the study is completely voluntary. Your decision not to consent will have no adverse consequences. Should you wish to withdraw during the experiment you can simply quit the webpage. All incomplete responses will be deleted. After you have completed the study and submitted your responses, it will no longer be possible to withdraw from the study, as your data will not be identifiable and able to linked to you.

\noindent \textbf{5. Possible benefits and risks to participants}

\noindent By participating in this study you will be contributing to research related to understanding what kinds of explanations are useful to people who use or who are impacted by automated fact checking systems. This is a long-term research project, so the benefits of the research may not be seen for several years. It is not expected that taking part will cause any risk, inconvenience or discomfort to you or others.

\noindent \textbf{6. What personal data does the project process?}

\noindent The project does not process any personal data.

\noindent \textbf{7. Participants’ rights under the General Data Protection Regulation (GDPR)}

\noindent As a participant in a research project, you have a number of rights under the GDPR. Your rights are specified in 
[anonymised]

\noindent \textbf{8. Person responsible for storing and processing of data}

\noindent[anonymised]



\noindent Please click 'Next' to read more about consenting to participate in the study.

\subsection{Human Evaluation Consent Form}
\label{app:consentform}

We hereby request your consent for processing your data. We do so in compliance with the General Data Protection Regulation (GDPR).
See the information sheet on the previous screen for more details about the project and the processing of your data.

\begin{itemize}[leftmargin=*]
    \item I confirm that I have read the information sheet and that this forms the basis on which I consent to the processing of my data by the project.

    \item I hereby give my consent that  [anonymised]
    may register and process my data 

    \item I understand that any data I provide will be anonymous and not identifiable to me.

    \item I understand that my anonymous response data will be retained by the study team.

    \item I understand that after I submit my responses at the end of the study, they cannot be destroyed, withdrawn, or recalled, because they cannot be linked with me.

    \item I understand that there are no direct benefits to me from participating in this study

    \item I understand that anonymous data shared through publications or presentations will be accessible to researchers and members of the public anywhere in the world, not just the EU.

    \item I give my consent that the anonymous data I provided may be stored in a database for new research projects after the end of this project.

    \item I give permission for my anonymous data to be stored for possible future research related to the current study without further consent being required.

    \item I understand I will not be paid for any future use of my data or products derived from it.
\end{itemize}

\noindent By checking this box, I confirm that I agree to the above and consent to take part in this study.

$\Box$ I consent

\subsection{Evaluation Task Instructions}
\label{app:instructs}

\textbf{What do I have to do?}\\
In this study you will see claims, an AI system’s prediction about whether this claim is true or false, how certain the system is about its label, and the corresponding evidence used to make the prediction. You will also see three different explanations for why the AI system is certain or uncertain about its prediction. These explanations are intended to help you decide how to interpret the true/false prediction. \\
Your task is to \textbf{evaluate the quality of the explanations} provided, \textbf{not} the credibility of the claims and evidence.
\\

\noindent \textbf{What information will I be shown?}\\
You will be shown examples of claims, evidence document, verdicts and explanations.

\begin{itemize}[leftmargin=*]
    \item A claim is some statement about the world. It may be true, false, or somewhere in between.
    \item Additional information is typically necessary to verify the truthfulness of a claim - this is referred to as evidence or evidence document. An evidence document consists of one or several sentences extracted from an external source for the particular claim. In this study, you will see two evidence documents that have been retrieved for a claim. These evidence documents may or may not agree with each other.
    \item Based on the available evidence, a verdict is reached regarding whether a claim is true or false.
    \item Uncertainty often arises when evaluating the claim and evidence to reach a verdict. Each verdict is accompanied by a numerical uncertainty score which represents the AI system's confidence that its predicted verdict is correct.
    \item You will see 3 alternative explanations for where uncertainty arises with regard to the verdict. Note that these explanations focus on the AI system's uncertainty, not the verdict itself.

    \item You are asked to evaluate the explanations according to 5 different properties. The properties are as follows: 

    \noindent\textbf{Helpfulness.} The explanation contains information that is helpful for evaluating the claim and the fact check.

    \noindent\textbf{Coverage. }The explanation contains important, salient information and does not miss any important points that contribute to the fact check. 

    \noindent\textbf{Non-redundancy. }The explanation does not contain any information that is redundant/repeated/not relevant to the claim and the fact check. 
    
    \noindent\textbf{Consistency.} The explanation does not contain any pieces of information that are contradictory to the claim and the fact check. 
    
    \noindent\textbf{Overall Quality. }Rank the explanations by their overall quality. 
    
    \noindent\item Please rank the explanations in descending order. For example, you should rank the explanation that you think is most helpful as ‘1’, and the explanation that you think is least helpful as ‘3’. If two explanations appear almost identical, you can assign them the same ranking, but as a general rule, you should try rank them in hierarchical order.

    \noindent\item The three explanations, Explanation A, Explanation B, and Explanation C, will appear in a different order throughout the study, so you may need to pay some attention to which is which.
\end{itemize}

\textbf{Important:}
Please only consider the provided information (claim, evidence documents, and explanations) when evaluating explanations. Sometimes you will be familiar with the claim, but we ask you to approach each claim as new, whether or not you have seen it before. It doesn’t matter whether you personally agree or disagree with the claim or evidence - we are asking you to evaluate what the AI produces: if you were to see this claim for the first time, would you find the explanation provided by the AI useful?
On the next page, you will see an example of the task.

\subsection{Example of human evaluation set-up}
\label{app:example_annotation}

Here is an example of what you will see during the study. First, you will see a \textbf{Claim}, and two pieces of \textbf{Evidence}, along with an AI system's predicted \textbf{Verdict} and the system's \textbf{Certainty} that its prediction is correct.

The \textbf{parts of the claim and evidence that are most important to the AI system's certainty are highlighted.} Parts of the Claim are Red, parts of Evidence 1 are Blue, and parts of Evidence 2 are Green.

Underneath, you will see \textbf{three alternative explanations for the AI system's certainty}, Explanation A, Explanation B, and Explanation C. The parts of each explanation that refer to the claim and evidence are colour coded in the same way (Claim = Red, Evidence 1 = Blue, Evidence 3 = Green).

Your task is to read the claim, evidence, and explanations, and rank each explanation based on five properties.

Now, you can try this example below!

\end{document}